\documentclass[conference]{IEEEtran}

\usepackage{multicol}
\usepackage[bookmarks=true]{hyperref}
\usepackage{amsmath}
\usepackage{amssymb}
\usepackage{graphicx}
\usepackage{booktabs}
\usepackage{subcaption}
\usepackage{color}
\usepackage{cite}
\usepackage{xcolor}

\usepackage[font=small]{caption} %
\usepackage[T1]{fontenc}

\title{REASSEMBLE: A Multimodal Dataset for Contact-rich Robotic Assembly and Disassembly}
\author{
        \IEEEauthorblockN{Daniel Sliwowski\IEEEauthorrefmark{1},
        Shail Jadav\IEEEauthorrefmark{1},
        Sergej Stanovcic\IEEEauthorrefmark{1},
        Jędrzej Orbik\IEEEauthorrefmark{1},
        Johannes Heidersberger\IEEEauthorrefmark{1},
        Dongheui Lee\IEEEauthorrefmark{1}\IEEEauthorrefmark{2}
        }
        \IEEEauthorblockA{\IEEEauthorrefmark{1}Autonomous Systems Lab, Institute of Computer Technology, TU Wien, Vienna, Austria.}
        \IEEEauthorblockA{\IEEEauthorrefmark{2}Institute of Robotics and Mechatronics (DLR), German Aerospace Center, Wessling, Germany.}
        Email: \texttt{\{\href{mailto:daniel.sliwowski@tuwien.ac.at}{daniel.sliwowski},
                         \href{mailto:shail.jadav@tuwien.ac.at}{shail.jadav},
                         \href{mailto:sergej.stanovcic@tuwien.ac.at}{sergej.stanovcic},} \\
                         \texttt{
                         \href{mailto:jedrzej.orbik@tuwien.ac.at}{jedrzej.orbik},
                         \href{mailto:johannes.heidersberger@tuwien.ac.at}{johannes.heidersberger},
                         \href{mailto:dongheui.lee@tuwien.ac.at}{dongheui.lee}\}@tuwien.ac.at}
    }

\begin{document}

\makeatletter
\let\@oldmaketitle\@maketitle%
\renewcommand{\@maketitle}{\@oldmaketitle%
  \begin{center}
        \captionsetup{type=figure}
        \includegraphics[width=\textwidth]{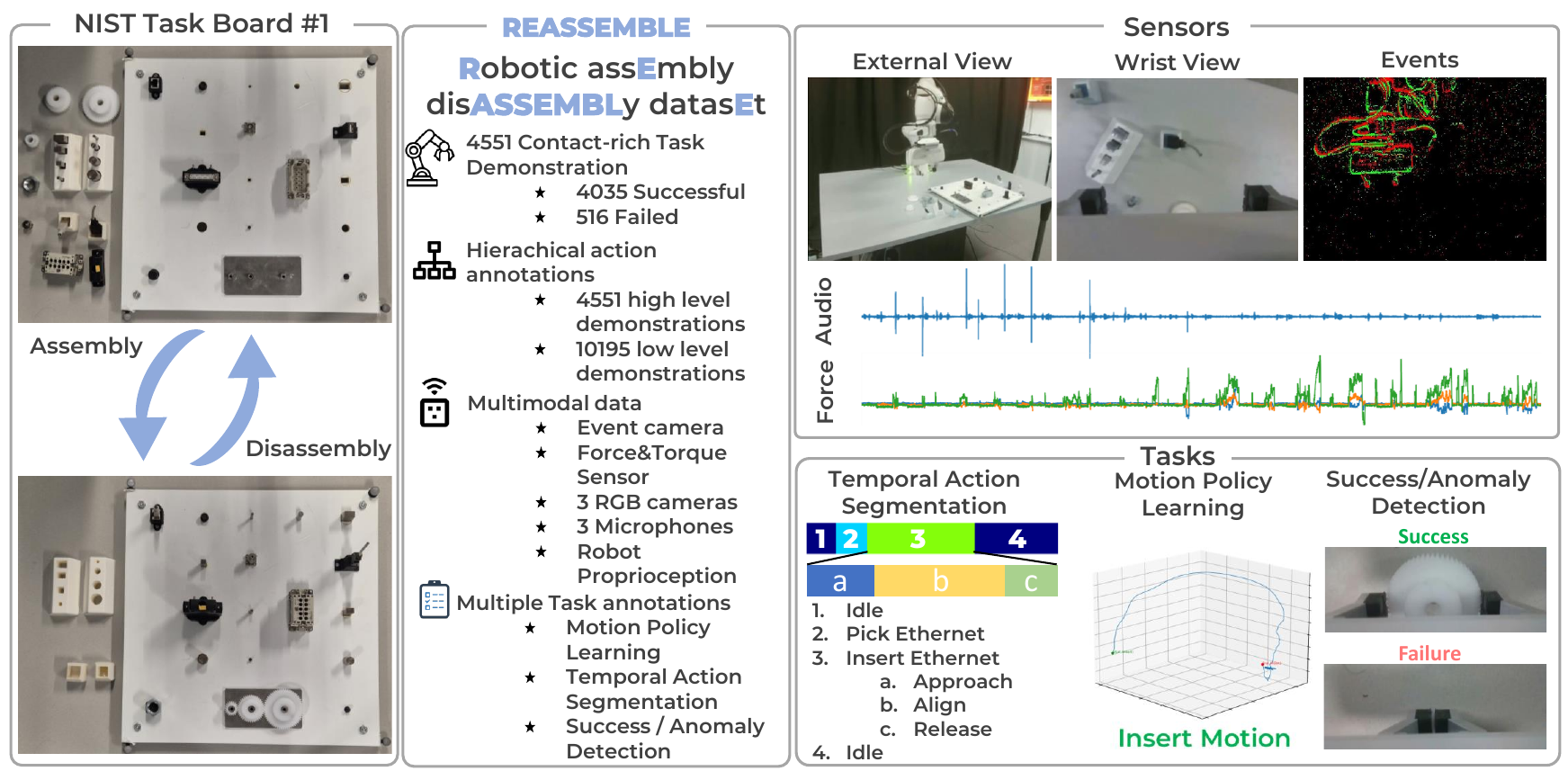}
        \captionof{figure}{\textbf{Overview of the REASSEMBLE dataset.} In REASSEMBLE, we focus on creating a dataset for contact-rich manipulation tasks. We leverage the well-established NIST Assembly Task Board \#1 \cite{kimble2022performance} to facilitate deployment of learned algorithms across different research institutes. The dataset includes data from various modalities, such as RGB cameras and robot proprioception, which are common in other works. Additionally, we incorporate event cameras, a force and torque sensor, and microphones, which are less common in manipulation datasets, and which will be beneficial for the community. We annotate the data for three different tasks: Hierarchical Temporal Action Segmentation (high-level actions and low-level skills), Motion Policy Learning, and Success/Anomaly Detection.} 
        \label{fig:teaser}
    \end{center}%
}
\makeatother

\maketitle
\addtocounter{figure}{-1}

\begin{abstract}
Robotic manipulation remains a core challenge in robotics, particularly for contact-rich tasks such as industrial assembly and disassembly. Existing datasets have significantly advanced learning in manipulation but are primarily focused on simpler tasks like object rearrangement, falling short of capturing the complexity and physical dynamics involved in assembly and disassembly. To bridge this gap, we present REASSEMBLE (Robotic assEmbly disASSEMBLy datasEt), a new dataset designed specifically for contact-rich manipulation tasks. Built around the NIST Assembly Task Board 1 benchmark, REASSEMBLE includes four actions (pick, insert, remove, and place) involving 17 objects. The dataset contains 4,551 demonstrations, of which 4,035 were successful, spanning a total of 781 minutes. Our dataset features multi-modal sensor data, including event cameras, force-torque sensors, microphones, and multi-view RGB cameras. This diverse dataset supports research in areas such as learning contact-rich manipulation, task condition identification, action segmentation, and task inversion learning. The  REASSEMBLE will be a valuable resource for advancing robotic manipulation in complex, real-world scenarios. The dataset is publicly available on our \href{https://tuwien-asl.github.io/REASSEMBLE_page/}{project website\footnote{https://tuwien-asl.github.io/REASSEMBLE\_page/}}.
\end{abstract}%

\IEEEpeerreviewmaketitle

\section{Introduction}
Robot learning, and deep learning more broadly, has made remarkable progress in recent years, driven by the increasing availability of high-quality, large-scale datasets~\cite{Yaprcr2019deeplearning}. These datasets have enabled state-of-the-art algorithms to achieve impressive performance across diverse tasks, including motion planning~\cite{chi2023diffusion, Brohan-RSS-23, zitkovich2023rt}, task planning~\cite{saycan2022arxiv, rana2023sayplan, driess2023palm}, locomotion~\cite{RoboImitationPeng20}, and anomaly detection~\cite{saycan2022arxiv, Sinha-RSS-24}. 

Most previous work in robot learning has focused on relatively simple manipulation tasks, such as object rearrangement, picking, and placing, which typically involve short horizons and limited interaction. In contrast, long-horizon, contact-rich manipulation tasks, such as object assembly and disassembly, remain underexplored. These tasks demand a deeper understanding of interaction dynamics and the ability to plan and execute precise, goal-oriented sequences. Recent work~\cite{wu2024tacdiffusion} has shown that current algorithms struggle with such tasks, largely due to the lack of datasets tailored for long-horizon, contact-rich scenarios. Without suitable data, models fail to generalize effectively in these challenging settings.

While industrial environments mitigate these challenges through specialized tooling and structure, robots in homes or unstructured settings face greater complexity. Tasks often involve diverse objects, irregular conditions, and unexpected events. For example, a household robot assembling furniture must handle uncertainties in alignment, material tolerances, and force application, without the aid of precise jigs or fixtures. In such cases, small deviations can easily lead to task failure. \textcolor{black}{Many approaches tackle these challenges by decomposing complex actions into simpler skills that can be sequenced to perform long-horizon tasks~\cite{sun2024hierarchical, pertsch2021accelerating}. For instance, fastening a screw can be broken down into approaching the socket, aligning the screw, and twisting it until secured. This hierarchical learning improves generalization, enabling the reuse of previously learned skills to solve new, unseen tasks~\cite{pertsch2021accelerating}.}

Although much of the focus has traditionally been on enabling robots to assemble products, there exists an equally critical need to develop robots capable of disassembling objects. Disassembly is a cornerstone of sustainability, particularly in advancing the circular economy. According to the United Nations Sustainable Development Goal 12, global demand for natural resources is projected to triple by 2050, far exceeding what the planet can sustainably provide under current consumption patterns. This underscores the urgent need for innovative solutions to improve resource efficiency and increase recycling rates. Robots capable of disassembling objects can play a transformative role in this effort. For instance, they could be deployed to efficiently disassemble complex objects such as cars, electronic devices, or EV batteries, which are currently dismantled manually in labor-intensive and hazardous processes. Despite the critical importance of disassembly in recycling, automation in this sector remains limited. \textcolor{black}{Additionally, this disassembly can be facilitated through task inversion learning (TIL) using assembly demonstrations.} Developing robust, adaptive agents capable of safely and efficiently disassembling products could drastically reduce waste, conserve resources, and mitigate the environmental impact of human consumption.

To bridge the gap between these pressing challenges, we introduce REASSEMBLE, a comprehensive dataset tailored to long-horizon and contact-rich manipulation tasks. While it draws inspiration from industrial assembly and disassembly, REASSEMBLE’s impact extends beyond industrial settings, offering opportunities to advance robotic capabilities for unstructured and everyday environments. Built upon the standardized NIST Assembly Task Boards~\cite{kimble2022performance}, specifically Task Board \#1, REASSEMBLE provides benchmarks for tasks such as gear meshing, peg insertions, electrical connector assemblies, and fastening nuts, as shown in Figure \ref{fig:teaser}. These tasks mimic the intricate manipulation challenges that robots face in real-world applications. What sets REASSEMBLE apart from other robot manipulation datasets is its focus on multimodal data for holistic learning frameworks. We provide a comparison of commonly used robot learning datasets and their properties in Table~\ref{tab:datasets}. The REASSEMBLE dataset includes multi-view RGB images and robot proprioceptive data, such as joint states, gripper states, and end-effector poses. Additionally, it incorporates less commonly utilized sensory modalities, such as audio recordings from external and wrist-mounted microphones, as well as six-axis force-torque measurements. Importantly, REASSEMBLE also includes event camera data. Unlike standard RGB cameras, event cameras capture changes in light intensity at each pixel rather than recording entire frames. This low-latency, high-precision information provides critical insights into object movement, enabling models to focus attention effectively and improving robustness in dynamic environments. Furthermore, REASSEMBLE distinguishes itself by providing multi-task annotations, enabling applications across diverse fields such as \textcolor{black}{hierarchical} temporal action segmentation, motion policy learning, and anomaly detection. By offering a rich, multi-modal dataset, REASSEMBLE fosters the development of adaptive and versatile robotic systems capable of tackling the challenges of long-horizon, contact-rich manipulation. To summarize, our contributions are as follows:
\begin{enumerate}
    \item A multi-modal dataset designed for long-horizon, high-precision, contact-rich assembly and disassembly tasks.
    \item A dataset with multi-task labels to support algorithm development in various robot learning fields, like \textcolor{black}{hierarchical} temporal action segmentation, motion policy learning, and anomaly detection.
    \item Data and annotations supporting both forward and inverse manipulation task learning.
    \item Introduction of the, to the best of our knowledge, first manipulation-focused dataset to include event camera data, providing low-latency and precise object and robot motion information.
\end{enumerate}
\begin{table*}[t]
    \centering
    \begin{tabular}{lcp{1.2cm}p{3cm}p{2.5cm}p{3cm}p{2cm}}
    \toprule
    Dataset & \# Demos & \# Verbs & Sensors & Robot & Collection method & Tasks \\
    \midrule
    BC-Z~\cite{jang2021bc} & 263k & 8 & 1 RGB Camera, Robot Proprioception & Everyday Robots & VR teleoperation & MPL \\
    RT-1~\cite{Brohan-RSS-23} & 130k & 8 & 1 RGB Camera, Robot Proprioception & Everyday Robots & VR teleoperation & MPL \\
    Language Table & 594k & n/r & 1 RGB Camera, Robot Proprioception & XArm & VR teleoperation & MPL \\
    BridgeDatav2~\cite{walke2023bridgedata} & 60.1k & 14 & 4 RGB Cameras, 1 Depth Camera, Robot Proprioception & WidowX & VR teleoperation & MPL \\
    RoboSet & 98.5k & 14 & 4 RGBD Cameras, Robot Proprioception & Franka Emika Panda Reaserach 3 & Kinesthetic, VR teleoperation, Autonomous & MPL \\
    FurnitureBench~\cite{heo2023furniturebench} & 5.1k & 9 & 2 RGB Cameras, Robot Proprioception & Franka & VR teleoperation & MPL \\
    DROID~\cite{droid} & 76k & 86 & 3 RGBD CAmeras, Robot Proprioception & Franka Emika Panda Reaserach 3 & VR teleoperation & MPL \\
    RH20T~\cite{fang2024rh20t} & 110k & 42 &  8-10 RGBD Cameras, 1 Microphone, F/T Sensor & Multiple & haptic teleoperation & MPL \\
    \midrule
    JIGSAWS~\cite{gao2014jhu} & 103 & 11 & 1 RGB Camera, Robot Proprioception & DaVinci & Haptic Teleoperation & TAS \\
    50SALADS~\cite{50Salads} & 50 & 6 & 1 RGBD Camera, accelerometers & N/A & Human demonstration & TAS \\
    Assembly101~\cite{sener2022assembly101} & 4.3k & 24 & 8 RGB Cameras, 4 Mono Cameras & N/A & Human demonstration & TAS \\
    \midrule
    FALIURE~\cite{inceoglu2021fino} & 229 & 6 & 1 RGBD Camera, 1 Microphone & Baxter & Scripted & AD \\
    (Im)PerfectPour~\cite{sliwowski2024conditionnet} & 554 & 4 & 2 RGB Cameras & Franka Emika Panda Reaserach 3 & VR teleoperation & AD \\
    \midrule
    REASSEMBLE & 4k & \textcolor{black}{4 actions and 9 skills} & 3 RGB cameras, Robot Proprioception, \textbf{1 DAVIS Event camera}, \textbf{3 Microphones}, \textbf{F/T Sensor} & Franka Emika Panda Reaserach 3 & Haptic teleoperation & \textbf{TAS, MPL, AD, TIL} \\
    \bottomrule
    \end{tabular}
    \caption{\textbf{Datasets comparison.} We compare several commonly used datasets based on the number of demonstrations, the number of verbs they contain, the sensors used during data collection, the robotic platform, the data collection method, and the tasks that can be learned from the dataset. The tasks include TAS (Temporal Action Segmentation), MPL (Motion Policy Learning), AD (Anomaly Detection), and TIL (Task Inversion Learning). We use "n/r" to denote information that is not reported by the works and "N/A" for cases where the category is not applicable.}
    \label{tab:datasets}
\end{table*}

\section{Related Works}
In robotic manipulation, most simulated environments and datasets primarily focus on fundamental tasks such as picking, placing, in-hand manipulation, lifting, and stacking \cite{rtx, droid,shafiullah2023bringing,walke2023bridgedata}, as shown in Table \ref{tab:datasets}. While these tasks are essential for understanding basic robotic capabilities, they are typically limited to short-horizon primitive skills and fail to provide the critical data necessary for precise, contact-rich manipulation. In contrast, complex long-horizon tasks such as assembly and disassembly, which require stringent tolerances, demand accurate and high-resolution contact information, including force-torque data, that visual sensing alone cannot reliably capture. For instance, force-torque sensing plays a pivotal role in assembly tasks involving a NIST assembly board, where precise measurements are essential to detect and respond to contact events with higher accuracy.

Existing datasets, such as the Furniture Benchmark \cite{heo2023furniturebench}, have made progress in addressing long-horizon tasks like furniture assembly. However, they lack the high-quality force-torque data required for the tight tolerances demanded in these applications. Similarly, while datasets like RH20T include manipulation data with audio and force-torque sensor information, they do not address the long-horizon task sequences needed for complex assemblies \cite{fang2024rh20t}. Vision-based tactile datasets, such as SPARSH, provide valuable insights into contact-rich manipulation, but they too fall short in capturing the extended temporal dynamics and multi-modal sensing required for precise assembly and disassembly tasks~\cite{higuera2024sparsh}. To address these limitations, REASSEMBLE introduces a novel dataset incorporating high-resolution force-torque sensing specifically tailored for tight-tolerance, high-precision, and long-horizon tasks.

The increasing prevalence of automation in robotic manipulation tasks highlights the necessity of effective skill assessment, task monitoring, and summarization to enhance system performance and reliability. Temporal action segmentation (TAS) plays a critical role in achieving these objectives by identifying and classifying distinct actions within continuous temporal sequences. Numerous datasets have been developed to support temporal action segmentation~\cite{50Salads,GTEA,breakfast}. For example, the 50Salads dataset is a widely used benchmark where humans perform salad preparation tasks \cite{50Salads}. This dataset provides temporally labelled actions for long-duration videos, facilitating the training and evaluation of models for action segmentation. These models can learn to segment complex temporal sequences into meaningful sub-tasks. However, such datasets primarily focus on human activity and often lack relevance to robotic manipulation tasks. Robotics-specific datasets, such as RT-1, Bridge V2, and RoboSet, offer data tailored to specific robotic tasks \cite{brohan2022rt,walke2023bridgedata}. While these datasets are valuable for advancing task-specific learning, they are not designed for comprehensive temporal action segmentation across diverse manipulation contexts. Moreover, conventional action segmentation datasets like 50Salads predominantly rely on visual and temporal information, often overlooking critical multimodal data, such as force-torque measurements, which are essential for understanding robotic actions. Therefore, REASSEMBLE also addresses robotic action segmentation and incorporates multimodal data, including visual, force-torque, and temporal streams. 

Advancements in large vision-language models (LVLMs), such as PaLM-E, LLaVa, have shown impressive capabilities in reasoning about task conditions by leveraging visual information during robotic task execution \cite{driess2023palm,li2023m}. These models provide robust insights into system states, enabling robots to monitor and adapt to dynamic conditions. However, their significant computational requirements make them impractical for real-time applications, where lightweight and efficient models are essential for rapid decision-making. To address these challenges, ConditionNet was introduced as a focused resource for condition monitoring in robotic systems \cite{sliwowski2024conditionnet}. While ConditionNet dataset ((Im)PerfectPour) is not aimed at contact-rich manipulation and long-horizon task. Building on these limitations, REASSEMBLE is designed to address the gaps in existing resources. It includes conditioned labeled multi-modal data such as visual, force-torque, and temporal streams, enabling more comprehensive condition monitoring. Crucially, it incorporates failure data to train models that can effectively learn to detect, understand, and respond to failures in real time. 

\section{Methodology}

\subsection{REASSEMBLE Tasks}
The REASSEMBLE dataset is designed to address long-horizon and contact-rich robotic manipulation tasks. Instead of creating custom tasks, we leverage the well-established NIST Assembly Task Board benchmark~\cite{kimble2022performance}, using it as the basis for our dataset. This benchmark includes four distinct task boards: (1) peg and connector insertion tasks, (2) manipulation of flexible belts and chains, and (3–4) cable routing tasks. For the REASSEMBLE dataset, we focus on Assembly Task Board 1, as it is the most feasible for single-arm robotic systems. In future expansions we plan to incorporate data from the remaining boards.

The assembly instructions for these task boards are available on the official NIST Assembly Task Boards \href{https://www.nist.gov/el/intelligent-systems-division-73500/robotic-grasping-and-manipulation-assembly/assembly}{\textcolor{blue}{website}}. For this work, we made minor modifications to the board design to enhance compatibility with our experimental setup. Specifically, we replaced the acrylic base plate with matte white PVC, a more cost-effective alternative. We also redesigned the Ethernet and USB holders to better accommodate the plugs used in our experiments, ensuring the plugs remain within the original tolerances specified by the benchmark. To facilitate object manipulation, we 3D-printed custom holders for pegs, Ethernet connectors, and USB plugs, allowing these connectors to remain vertical when not inserted into the board. This modification simplifies the picking and insertion tasks, aligning better with the kinematics of the Franka robot, which is more suited for top-down operations than lateral manipulations. The 3D models for these modified components are made publicly available on the project website. Furthermore, reflective markers were attached to the board’s corners to enable accurate localization using a motion capture system. Figure~\ref{fig:teaser} illustrates both the assembled and disassembled configurations of the task boards.

The REASSEMBLE dataset encompasses both assembly and disassembly processes, with the associated actions defined as follows: for assembly, the actions are \textit{picking }and \textit{inserting}, whereas for disassembly, the actions are \textit{removing} and \textit{placing}. These four actions represent the minimal set required to comprehensively describe the assembly and disassembly tasks. The manual labelling of demonstrations by human operators necessitated limiting the action set to these four categories, thereby reducing cognitive load during annotation. The assembly board contains a total of 20 objects; however, due to constraints in the robot and teleoperation setup, the three nuts were excluded from the dataset (M12, M8, and M4). As a result, the dataset comprises 17 objects. A detailed breakdown of the objects included in the dataset is provided in \textcolor{black}{the Appendix in Figure~\ref{fig:objects}}.

\subsection{Sensors}
We focus on collecting a comprehensive range of sensory information during task demonstrations to create a robust dataset. Consistent with prior works, we capture multi-view RGB video using two external cameras and one wrist-mounted camera. The external cameras are HAMA C-600 Pro webcams, while the wrist-mounted camera is an Intel RealSense D435i. In addition, we record proprioceptive data from the robot, including joint positions, velocities, efforts, end-effector position and orientation (relative to the robot's base frame), and gripper width.

\begin{figure}
    \centering
    \includegraphics[width=\linewidth]{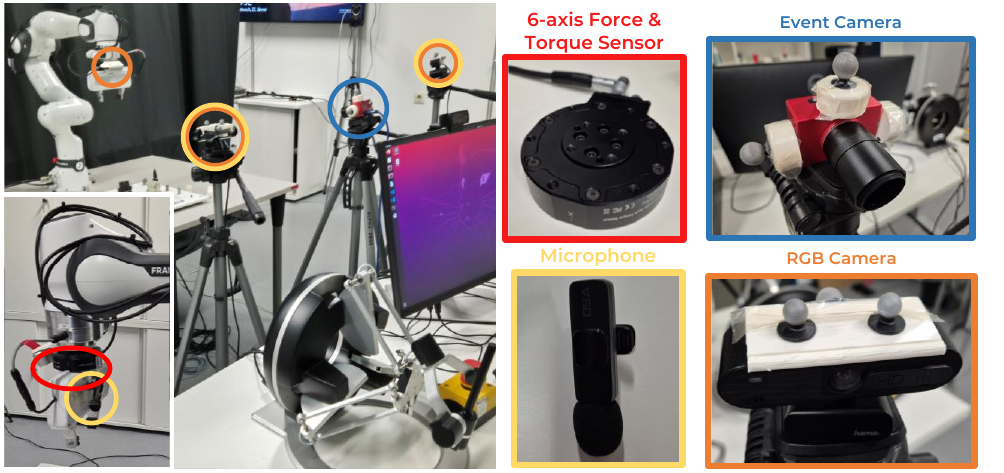}
    \caption{\textbf{Overview of the sensor placement.} We use two external and one wrist-mounted RGB cameras (marked in orange). Additionally, we use an externally mounted event camera (in blue), three microphones (in yellow), and one wrist-mounted force/torque (F/T) sensor (in red). The omega.6 haptic teleoperation device is also visible.}
    \label{fig:sensor_setup}
\end{figure}

 To capture audio data, we utilize three microphones: two mounted on the external cameras and one attached to the robot's gripper. The gripper microphone is an OSA K1T wireless microphone. Interaction forces and torques are measured using a wrist-mounted 6-axis force-torque (FT) sensor (AIDIN ROBOTICS AFT200-D80-C), as shown in Figure \ref{fig:sensor_setup}. For accurate FT readings in the robot's base frame, we perform a calibration procedure. This involves sampling 6 unique orientations uniformly distributed over a unit sphere, moving the robot to these orientations, and recording FT measurements. Using the sensor pose and these measurements, we calibrate the gripper’s mass, center-of-mass pose, and sensor bias, following the methodology in~\cite{Kubus2007calib}. The calibration process is implemented using the ${\tt force-torque-tools}$ ROS package.

In addition to the aforementioned sensors, the dataset includes an event camera. The sensor is mounted externally on a tripod in front of the robot. Due to its limited resolution and field-of-view, it is positioned at a distance that allows it to capture the majority of the workspace of the robot while still being able to perceive the smaller objects, such as small gears or pegs.

To estimate the position and orientation of the cameras relative to the robot's base frame, we employ a motion capture system. Custom 3D-printed brackets with attached reflective markers are mounted on each camera, ensuring precise localization. Reflective markers are also attached to the robot's base frame to simplify the computation of relative poses. An overview of the complete sensor setup is presented in Figure~\ref{fig:sensor_setup}.

\begin{figure}[htbp]
    \centering
    \begin{subfigure}{0.49\linewidth}
        \centering
        \includegraphics[width=\linewidth]{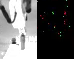}
        \caption{\textcolor{black}{Before nudge.}}
        \label{fig:before}
    \end{subfigure}
    \hfill
    \begin{subfigure}{0.49\linewidth}
        \centering
        \includegraphics[width=\linewidth]{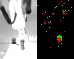}
        \caption{\textcolor{black}{After nudge.}}
        \label{fig:after}
    \end{subfigure}
    \caption{\textcolor{black}{Visualization of event camera data. In this example, a peg becomes stuck after insertion, and the robot applies a nudge to properly insert it. (a) shows a snapshot of the event stream before the nudge, and (b) after. The motion of the peg is clearly visible in the event camera stream.}}
    \label{fig:EC_image}
    \vspace{-0.1cm}
\end{figure}

 Event cameras are emerging as critical sensors in robotics, providing a fundamentally different data acquisition paradigm compared to traditional RGB cameras \cite{gehrig2024low}. Unlike frame-based cameras that capture static images at fixed intervals, event cameras asynchronously record changes in pixel intensity, resulting in high temporal resolution and low latency. Event cameras can provide valuable information for robotic manipulation by capturing motion events with exceptional precision. For instance, during manipulation tasks, event cameras can highlight fine-grained movements, such as the subtle shifts of objects in the environment. \textcolor{black}{An example of this is shown in Figure~\ref{fig:EC_image}, where the robot nudges a stuck peg. Once dislodged, the peg’s movement is clearly captured in the event data. Event camera information has also been successfully used in tasks such as slip detection~\cite{reinold2025combined}, and has shown improved performance in human action recognition compared to using only RGB images~\cite{chadha2019neuromorphic}. Motivated by these findings, we incorporated event camera data into REASSEMBLE's multi-modal framework.}

\subsection{Teleoperation Setup}

\begin{figure}[b]
    \centering
    \includegraphics[width=\linewidth]{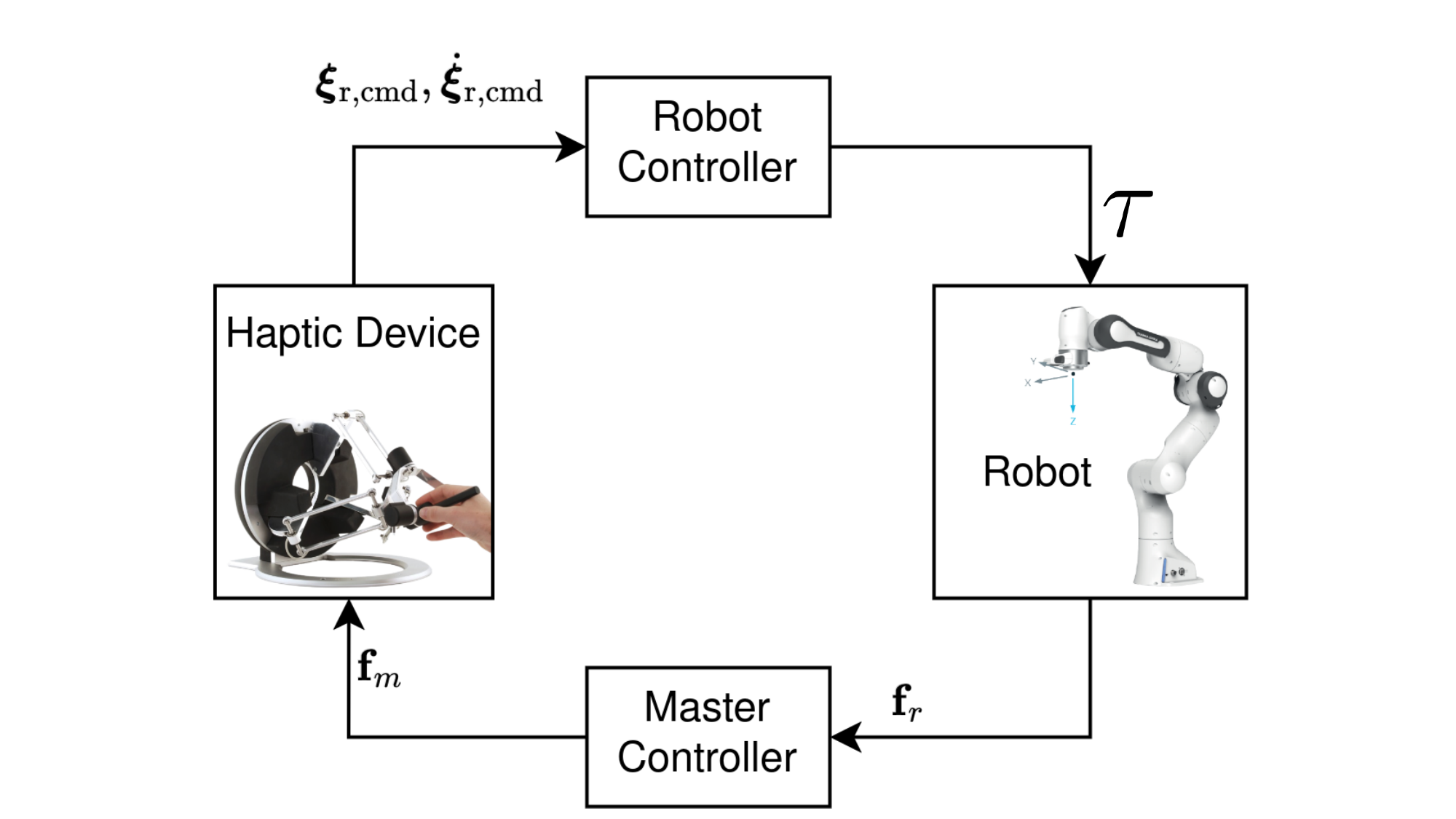}
    \caption{Overview of the teleoperation control system. The operator controls the robot's motion through the haptic device, which simultaneously feeds back forces measured at the robot's end effector.}
    \label{fig:teleoperation}
\end{figure}

During data collection, the operator controls the system via a haptic device. Haptic feedback is especially important for demonstrating contact-rich manipulation tasks, as the forces perceived at the end effector are critical. The haptic device allows the operator to feel the interaction forces with the environment and adjust the demonstrated motion accordingly. Without this feedback, the demonstrated forces could be incorrect; for example, pulling a plug too hard might cause the entire board to move, or exceeding the robot's force thresholds could damage the connectors during insertion. Haptic feedback also provides cues to the operator about the success of the tasks, such as correctly aligning a peg with a hole or properly attaching a nut to a bolt. The teleoperation system consists of a haptic device (omega.6 from Force Dimension) as the master and a Franka Emika FR3 robotic arm as the remote device. The system operates bidirectionally, allowing the master device to control the robot's end-effector pose while relaying force feedback from the robot back to the master device \cite{michel_bilateral_2021}, see Figure \ref{fig:teleoperation}. The haptic feedback allows the operator to perceive the robot's interaction forces with the environment, enabling them to adjust the motion accordingly.

To allow for intuitive user control and effectively utilize the robot's workspace, the haptic device's position \( \mathbf{x}_m \in \mathbb{R}^3 \) must be aligned with the robot's base frame and its motion appropriately scaled. This transformation is achieved through an affine operation that combines uniform scaling, represented by the diagonal matrix \( \mathbf{S} = \text{diag}(4.5, 4.5, 4.5) \in \mathbb{R}^{3 \times 3} \), and rotation alignment between the remote's and master's base frames using the rotation matrix \( \mathbf{R}_{\mathrm{m0}}^{\mathrm{r0}} \in \mathbb{R}^{3 \times 3} \).
The transformed master device position is then used to compute the command position for the robot's end effector:
\begin{align}
    \mathbf{x}_{\mathrm{r,cmd}} &= \mathbf{x}_r^{\text{init}} + \mathbf{R}_{\mathrm{m0}}^{\mathrm{r0}} \mathbf{S} (\mathbf{x}_m - \mathbf{x}_m^{\text{init}}),
\end{align}
where \( \mathbf{x}_r^{\text{init}} \in \mathbb{R}^3 \) is the initial position of the robot's end effector. The relative orientation change between the master's orientation \( \mathbf{q}_m = [ w_m, \mathbf{v}_m^T ] \in \mathbb{R}^4 \) and its initial orientation \( \mathbf{q}_m^{\text{init}} \in \mathbb{R}^4 \) is calculated as
\begin{align}
    \Delta \mathbf{q}_m = \mathbf{q}_m \otimes \mathbf{q}_m^{\text{init}^{-1}},
\end{align}
where \( \otimes \) denotes the quaternion product defined as
\begin{align}
    \mathbf{q}_1 \otimes \mathbf{q}_2 \triangleq \begin{bmatrix}
        w_1 + w_2 - \mathbf{v}_1^T \mathbf{v}_2\\
        w_2 \mathbf{v}_2 + w_2 \mathbf{v}_1 + \mathbf{v}_1 \times \mathbf{v}_2
    \end{bmatrix}.
\end{align}
Given this quaternion difference $\Delta \mathbf{q}_m$ and the robot's initial orientation $\mathbf{q}_s^{\text{init}}$, the target orientation of the robot end effector is computed as:
\begin{align}
    \mathbf{q}_{\mathrm{r,cmd}} &= \mathbf{q}_s^{\text{init}} \otimes \Delta \mathbf{q}_m.
\end{align}
\textcolor{black}{The control input \( \tau_{\rm{imp}} \in \mathbb{R}^7\) for the robot's Cartesian impedance controller is expressed as 
\begin{align}
\tau_{\rm{imp}} = \mathbf{J}^\top( \mathbf{K}_x \mathbf{e}_{\xi} + \mathbf{D} \mathbf{e}_{\dot{\xi}}),
\end{align}
where \( \mathbf{e}_{\xi} \in \mathbb{R}^6 \) is the pose error vector derived from \( \boldsymbol{\xi}_{\mathrm{r,cmd}} - \boldsymbol{\xi}_{\mathrm{r}} \), and \( \mathbf{e}_{\dot{\xi}} = \dot{\boldsymbol{\xi}}_{\mathrm{r,cmd}} - \dot{\boldsymbol{\xi}}_{\mathrm{r}} \in \mathbb{R}^6 \) is the twist error. 
Note that while \( \boldsymbol{\xi}_{\mathrm{r}} \in \mathbb{R}^7 \) and \( \boldsymbol{\xi}_{\mathrm{r,cmd}} \in \mathbb{R}^7 \) are the current and commanded Cartesian poses (with position \(\mathbf{x}_{\mathrm{r}} \in \mathbb{R}^3\) and quaternion orientation \(\mathbf{q}_{\mathrm{r}} \in \mathbb{R}^4\)), the pose error \( \mathbf{e}_{\xi} \) is computed as a 6D vector by appropriate transformation of the orientation difference. The matrices \( \mathbf{K}_x \in \mathbb{R}^{6 \times 6} \) and \( \mathbf{D} \in \mathbb{R}^{6 \times 6} \) represent the stiffness and damping coefficients, respectively.}

To provide the operator with haptic feedback, a six-axis force-torque sensor mounted on the robot's flange measures the external force \( \mathbf{f}_r \in \mathbb{R}^3 \), which is transformed from the sensor frame to the base frame of the haptic device
\begin{align}
    \mathbf{f}_r^{\mathrm{m0}} = \mathbf{R}_{\mathrm{m0}}^{\mathrm{r0}^{-1}} \mathbf{R}_{\mathrm{r,Sensor}}^{\mathrm{r0}} \mathbf{f}_r,
\end{align}
where \( \mathbf{R}_{\mathrm{r,Sensor}}^{\mathrm{r0}} \) is the rotation matrix aligning the sensor's frame with the robot's base frame. 
The force applied to the haptic device, denoted as \( \mathbf{f}_m \in \mathbb{R}^3 \), is computed as:
\begin{align}
    \mathbf{f}_m = - \mathbf{B}_m \dot{\mathbf{x}}_m + 0.35 \mathbf{f}_r^{\mathrm{r0}},
\end{align}
where \( \mathbf{B}_m = \text{diag}(40, 40, 40) \, \text{N}/(\text{m/s}) \) is the viscous damping matrix, and \( \dot{\mathbf{x}}_m \) is the velocity of the haptic device. The first term, \( - \mathbf{B}_m \dot{\mathbf{x}}_m \), provides stabilizing damping to suppress oscillations, while the second term scales and applies the processed force feedback, which allows the operator to perceive the interaction forces experienced by the robot's end effector in the remote environment.

For higher-DOF robots such as the FR3 robot, neglecting nullspace optimization can lead to undesirable configurations and complicate teleoperation. This is particularly problematic near kinematic singularities or when approaching joint limits. To address these issues, nullspace optimization is performed to ensure smooth operation. The optimization leads to joint torques \( \mathbf{\tau}_{\mathrm{null}} \) that do not affect the end-effector pose but optimize the nullspace configuration. These torques are computed to achieve two objectives: avoiding kinematic singularities and maintaining joint positions within safe limits. The nullspace torque, denoted by \(\boldsymbol{\tau}_{\mathrm{n}} \in \mathbb{R}^{7}\), is computed  as
\begin{align}
\boldsymbol{\tau}_{\mathrm{n}} = \left( \mathbf{I} - \mathbf{J}^\top(\boldsymbol{\theta}) \left[ \mathbf{J}(\boldsymbol{\theta}) \mathbf{J}^\top(\boldsymbol{\theta}) \right]^{-1} \mathbf{J}(\boldsymbol{\theta}) \right) \frac{\partial V(\boldsymbol{\theta})}{\partial \boldsymbol{\theta}}, 
\end{align} where \(\mathbf{J}(\boldsymbol{\theta}) \in \mathbb{R}^{6 \times 7}\) represents the robot's Jacobian matrix, and \(V(\boldsymbol{\theta}) \in \mathbb{R}\) is a potential function designed to incorporate multiple objectives. The objective function \(V(\boldsymbol{\theta})\) is defined as
\begin{align}
V(\boldsymbol{\theta}) = \alpha \det\left( \mathbf{J}(\boldsymbol{\theta}) \mathbf{J}^\top(\boldsymbol{\theta}) \right) + \beta \mathrm{J_{\rm L}}(\boldsymbol{\theta}) + \gamma \dot{\boldsymbol{\theta}},
\end{align} where \(\alpha, \beta, \gamma \in \mathbb{R}\) are weighting coefficients. The determinant term \(\det\left(\mathbf{J}(\boldsymbol{\theta}) \mathbf{J}^\top(\boldsymbol{\theta})\right)\) promotes the avoidance of singularities by maximizing the manipulability measure. The joint limit avoidance term \(\mathrm{J_{\rm L}}(\boldsymbol{\theta})\) is expressed as
\begin{align}
\mathrm{J_{\rm L}}(\boldsymbol{\theta}) = \tanh\left( a(\boldsymbol{\theta} - \mathbf{c}) + b \right) + \tanh\left( -a(\boldsymbol{\theta} - \mathbf{c}) + b \right) + 2,
\end{align}where \(a, b < 0 \in \mathbb{R}\) and \(\mathbf{c} \in \mathbb{R}^{7}\) are parameters that define the joint limit boundaries and scaling. The velocity damping term \(\gamma \dot{\boldsymbol{\theta}} \in \mathbb{R}^{7}\) penalizes high joint velocities, thereby ensuring smooth motion. By integrating the nullspace optimization torque \(\boldsymbol{\tau}_{\mathrm{n}}\), the teleoperation system effectively avoids undesirable configurations, respects joint limits, and maintains high levels of manipulability. These enhancements collectively provide the operator with a more intuitive and reliable teleoperation experience.

\subsection{Data Collection}
Each demonstration begins by randomizing the position and orientation of the board and objects within the robot's workspace. The operator is then instructed to assemble and disassemble the board by teleoperating the robot. During data collection, the operator simultaneously labels the actions by verbally narrating their activity and marks the temporal boundaries of task segments by pressing the appropriate key on the keyboard. To convert the audio file containing the narrated action descriptions, we use the Whisper~\cite{radford2023robust} automatic speech recognition algorithm. At the beginning of each demonstration, we measure the poses of the board and cameras using the motion capture system. We do not continuously track the board's pose because the event camera erroneously reports events when the motion capture system's illumination is turned on.

During teleoperation, all sensory information is recorded using the {\tt rosbag} tool. Subsequently, the raw recordings are post-processed into HDF5 files, \textcolor{black}{whose structure can be found in the Appendix.} We do not downsample or synchronize any sensory data during postprocessing; instead, we save the raw data along with timestamps. This approach allows for later adjustments in synchronization or data downsampling for specific downstream tasks. To optimize memory usage, all video and audio recordings are stored as encoded MP4 and MP3 files, respectively, which reduces the dataset size significantly, by a factor of six compared to the unprocessed rosbag recordings.

The accuracy of the predicted text for action narration depends on how clearly the operator spoke during data collection. We observed that some errors occur in the transcriptions; for example, "D-SUB" is often recognized as "the sub." Such errors can be easily corrected automatically by substituting incorrect phrases with the correct ones. Additionally, we manually validate the data stored in the recordings and the action segment and success annotations. For this, we develop and leverage a data visualization tool \textcolor{black}{which we describe in more detail in the Appendix}. We verify that all data sources cover the entire recording, are synchronized, and that the action and success labels accurately reflect the events in the videos. If any labelling errors are identified, we correct them. Additionally, if any data is missing, we record the corresponding file name. Finally, based on the known sets of objects and actions, we can compute all valid annotations. We leverage this information to validate whether all action annotations are correct and ensure that no transcription mistakes remain in the dataset.

\textcolor{black}{
The skill annotations were manually added after recording all demonstrations. We identified nine commonly shared skills across high-level actions. \textbf{Grasp} refers to free-space movement of the empty gripper until it closes on the object, followed by \textbf{Lift}, which involves moving the object upward. \textbf{Approach} denotes the free-space movement with an object inside of the gripper until the object makes contact with the socket, while \textbf{Align} describes a search motion to correctly position the object with respect to the socket. \textbf{Release} involves letting go of the object, either after alignment or when placing it on the table. \textbf{Twist} is used when rotating a BNC connector or nut, regardless of direction. \textbf{Push} refers to inserting the object into a socket, commonly seen with gears and waterproof connectors, and \textbf{Pull} denotes extracting the object from a socket. Finally, \textbf{Nudge} captures a gentle push to dislodge a peg that may be stuck after release. After completing the annotations, we verified that all skill names were free of typos and manually reviewed infrequent skill-action combinations to ensure they were correctly labeled and not erroneous.
}

\begin{figure}
    \centering
    \includegraphics[width=\linewidth]{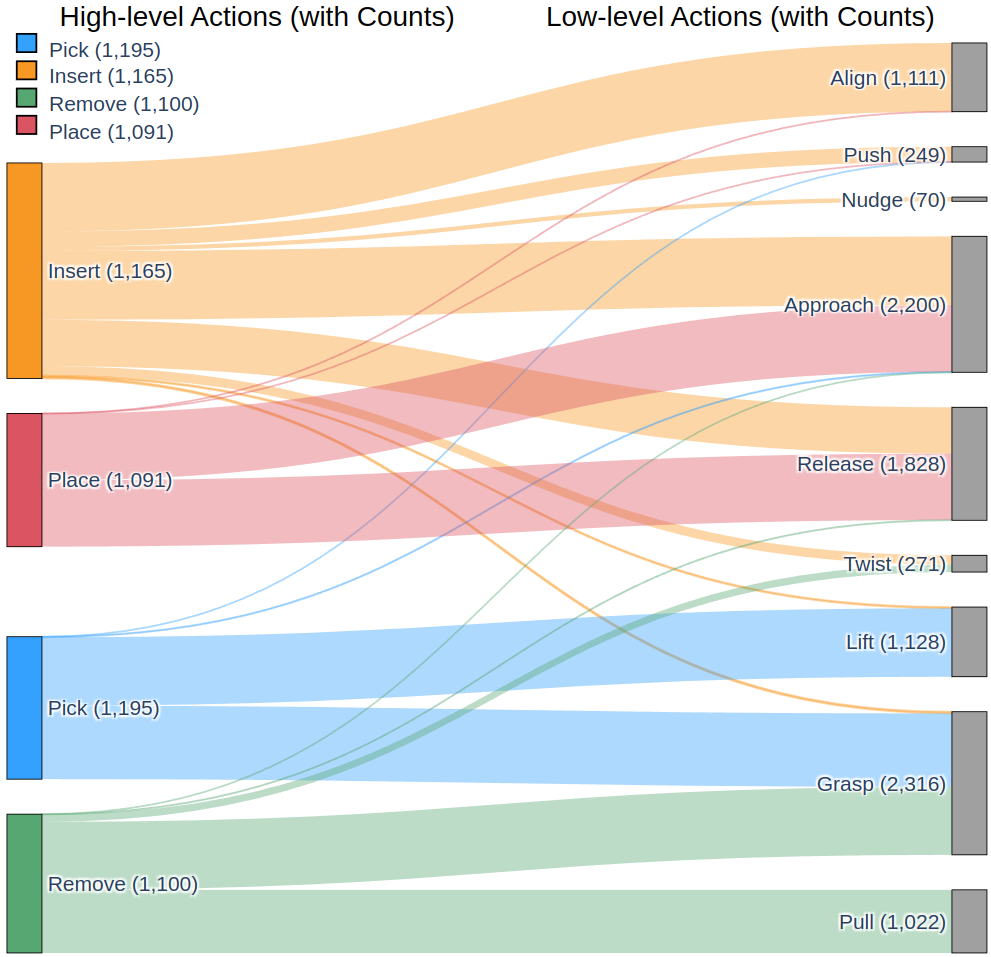}
    \caption{Sankey diagram showing the hierarchical structure and how skills are distributed within actions. The REASSEMBLE dataset contains 121 unique skill-object pairs.}
    \label{fig:hierarchy_sankey}
\end{figure}

\begin{figure*}
    \centering
    \begin{subfigure}[b]{0.79\textwidth}
        \includegraphics[trim=1.5cm 0cm 19cm 0cm, clip, width=\textwidth]{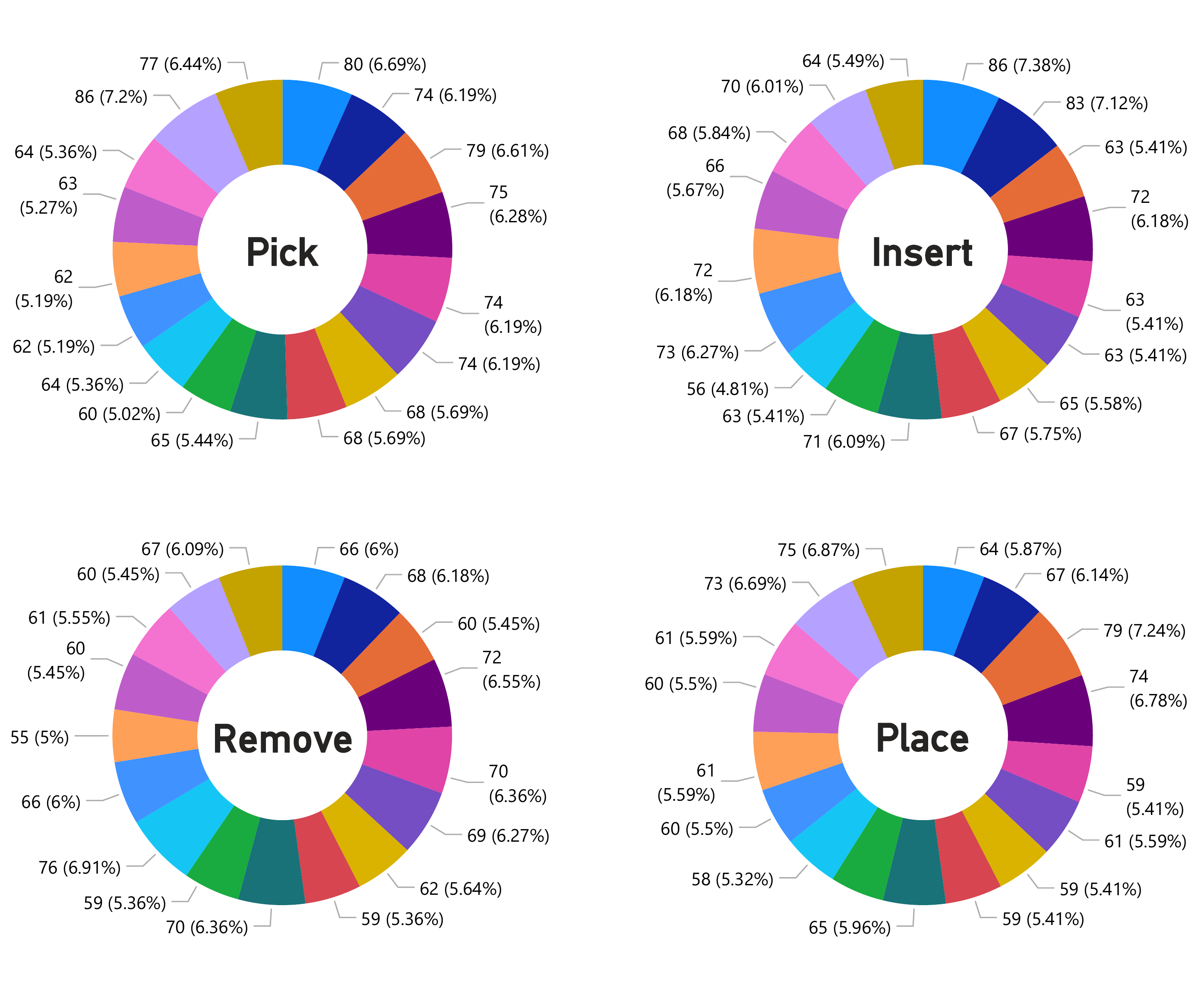}
    \end{subfigure}
    \begin{subfigure}[b]{0.19\textwidth}
        \includegraphics[trim=35cm 0cm 5cm 0cm, clip, height=12cm, keepaspectratio]{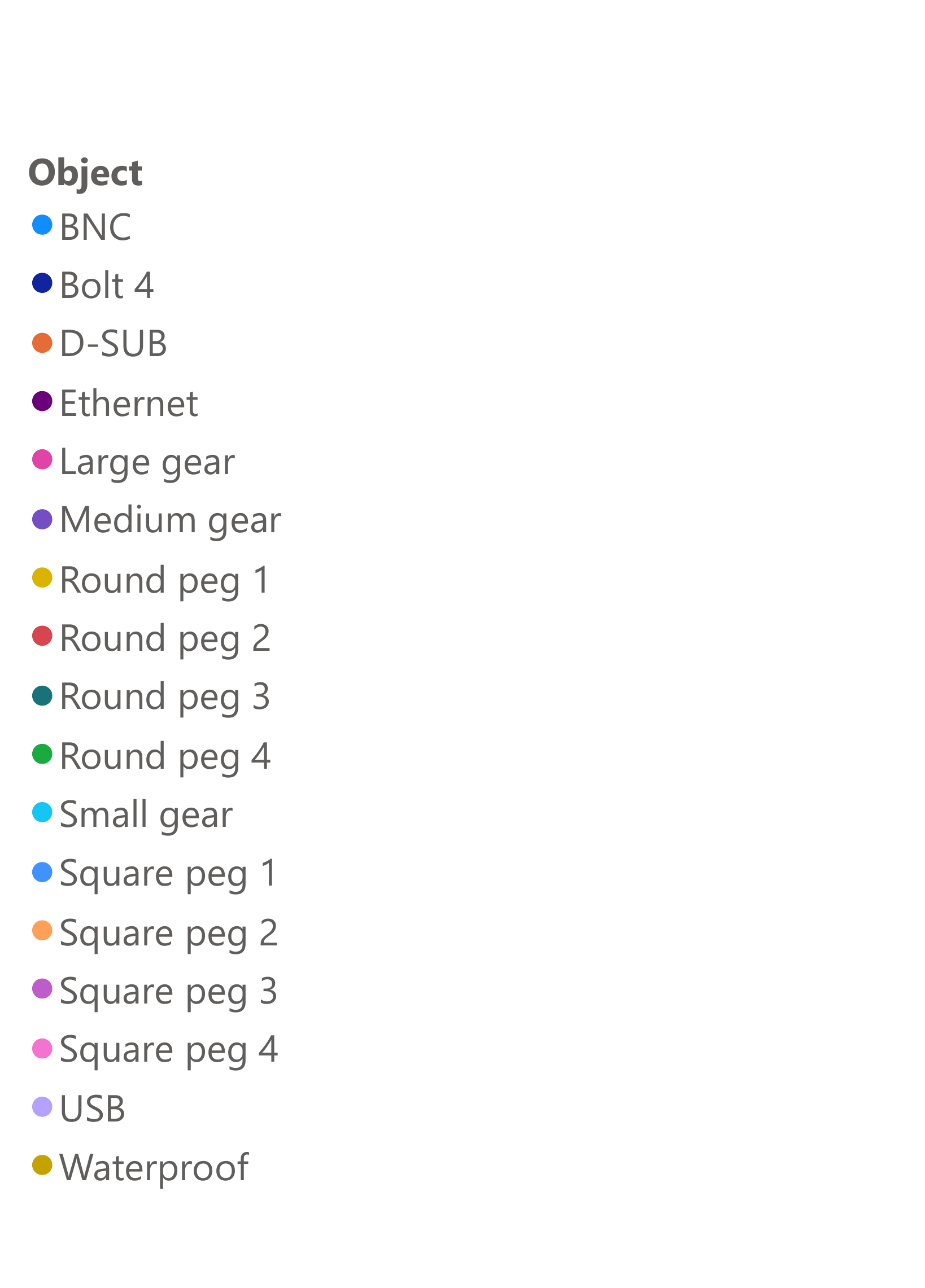}
    \end{subfigure}
    
    \caption{\textbf{Number of demonstrations of each action-object pair.} In REASSEMBLE, we have 4 actions: pick, insert, remove, and place, and 17 objects, resulting in 68 unique action-object pairs. The number of executions of each unique action is almost equal, making it a balanced dataset.}
    \label{fig:stats_ocur}
\end{figure*}

\begin{figure*}
    \centering
    \includegraphics[width=\linewidth,]{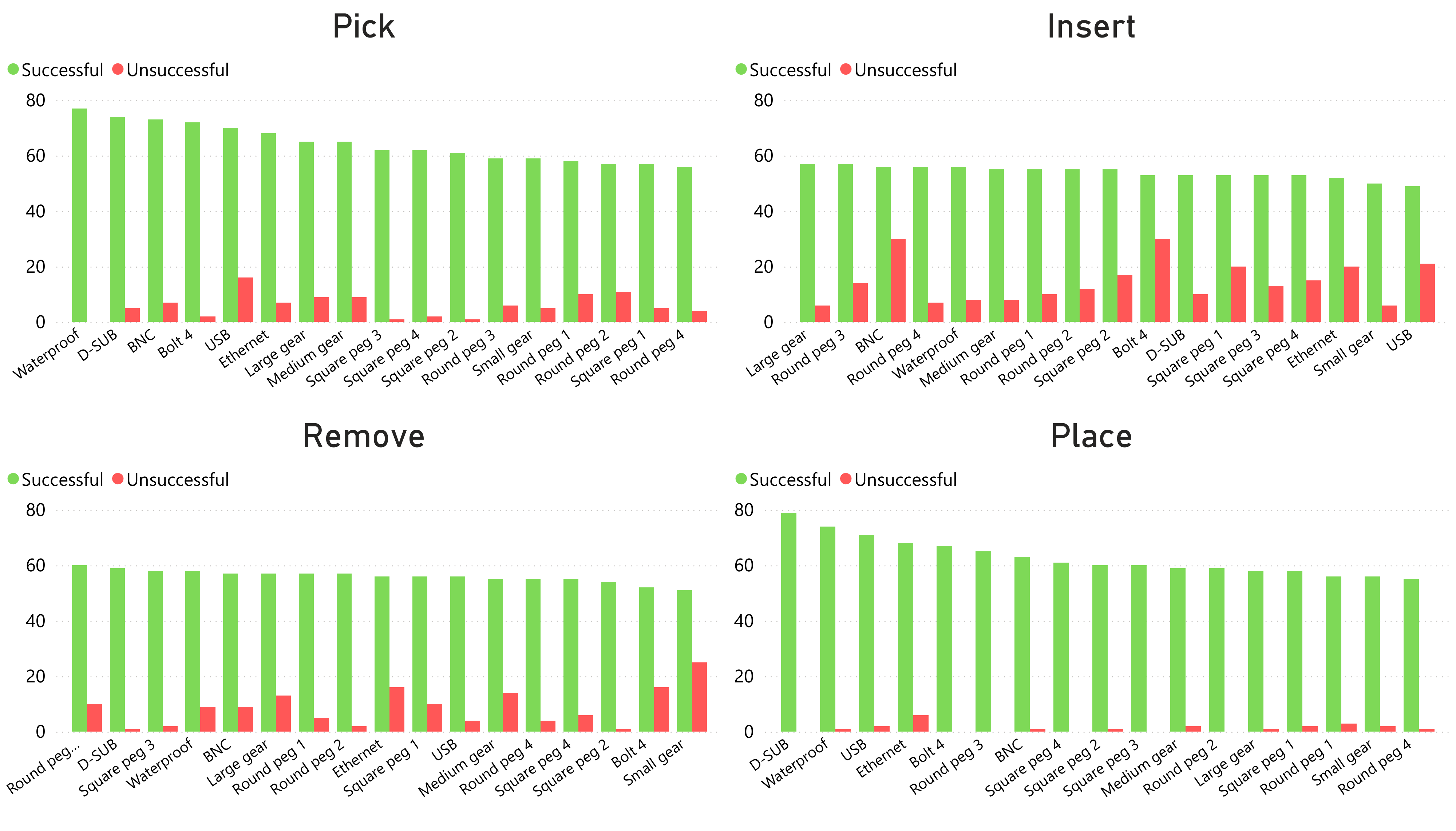}
    \caption{\textbf{Number of successful and unsuccessful demonstrations for each action-object pair.} In REASSEMBLE, we also annotate the success of each action demonstration. This can serve as a proxy for the action difficulty. The insert action is the most difficult, as it requires the highest precision and longest horizon of all actions, followed by the remove, pick, and place actions.}
    \label{fig:stats_succ}
\end{figure*}

\section{Dataset Statistics}
In this section, we conduct data analysis on the demonstrations contained within the REASSEMBLE dataset. We focus on four key aspects of the data: the number and distribution of action occurrences, the amount of successful and unsuccessful action demonstrations for each action, the diversity of positions where the robot interacts with the objects, and the patterns in the demonstrated force-torque profiles.

\subsection{Action distribution}
The number of demonstrations for each action-object pair and their distribution provide insights into the diversity and balance of the REASSEMBLE dataset. To achieve the best performance across all actions, it is crucial that the dataset contains a large and relatively equal number of demonstrations for each action. This ensures that downstream models observe each unique action an equal number of times, reducing performance bias toward more frequently occurring actions. The REASSEMBLE dataset includes four actions and 17 different objects, resulting in a total of 68 unique action-object pairs, or 69 when including the "Idle" action. Figure~\ref{fig:stats_ocur} shows the number of occurrences for each unique action. For clarity, we present one plot per verb in the REASSEMBLE dataset. 

The REASSEMBLE dataset contains a minimum of 55 demonstrations for the "Remove square peg 2" action and a maximum of 86 demonstrations for the "Pick USB" and "Insert BNC" actions. This demonstrates that the dataset is balanced.

\textcolor{black}{Figure~\ref{fig:hierarchy_sankey} illustrates the skill composition of each action. The dataset contains a total of 4,551 actions, decomposed into 10,195 skills with an average of 2.2 skills per action. On average, the \textit{Pick} action consists of three skills, while most other actions typically involve two. The \textit{Insert} action, which is the most challenging in the REASSEMBLE dataset, shows the greatest variability, with some demonstrations requiring up to ten steps. Insertion sequences with more than three skills often involve objects like the BNC connector or the nut. For instance, screwing the nut requires releasing it, rotating the gripper back, and re-grasping it, due to the limited rotational range of the robot’s final joint. Uncommon action-skill combinations typically occur during failure recovery. For example, a \textit{Push} during the \textit{Place} action appears once, when the Ethernet connector gets stuck in the gripper. Similarly, an \textit{Align} skill appears during \textit{Place} when a peg needed to be reinserted into its holder.
}

\subsection{Action difficulty and failure modes}
The number of failed demonstrations per action can serve as a metric for task difficulty, as operators are more likely to fail when the motion is complex. Figure~\ref{fig:stats_succ} shows the relationship between successful and unsuccessful demonstrations for each unique action. From the figure, we observe that the most difficult action in the dataset is the "Insert" action, which has the highest number of total failures (Figure~\ref{fig:stats_succ}, top right). This is due to the complexity of the insertion motion, which typically involves multiple steps: approaching the target, executing a search pattern to align the object with its socket, and finally applying downward force to insert the object. 

Among all insertion tasks, inserting the BNC and nut 4 are the most difficult. These tasks not only require aligning the object with the socket but also rotating it around the normal axis to the board to secure the object. The higher failure rate for inserting objects like the Ethernet cable or USB cable arises for two reasons: first, both plugs are directional and must have a specific orientation relative to the socket; second, the plugs are located on the edge of Task Board \#1, increasing the likelihood of being beyond the robot's workspace.

The majority of failures in the "Pick" action (Figure~\ref{fig:stats_succ}, top left) occur because the gripper either misses the object or the object slips out of the gripper while it is closing. Failures in the "Remove" action (Figure~\ref{fig:stats_succ}, bottom left) often result from improper alignment of the gripper with the object during removal, causing the object to "jam" in the socket and not be released. 

In contrast, we observe the fewest failures for the "Place" action (Figure~\ref{fig:stats_succ}, bottom right). The only requirement is that the object remains on the table after the action. However, failures in this action do occur if the object slips prematurely from the gripper and lands on the task board, which we classify as a failure in this work, as an object on the board might obstruct the sockets of other objects.

\begin{figure}
    \centering
    \includegraphics[width=\linewidth]{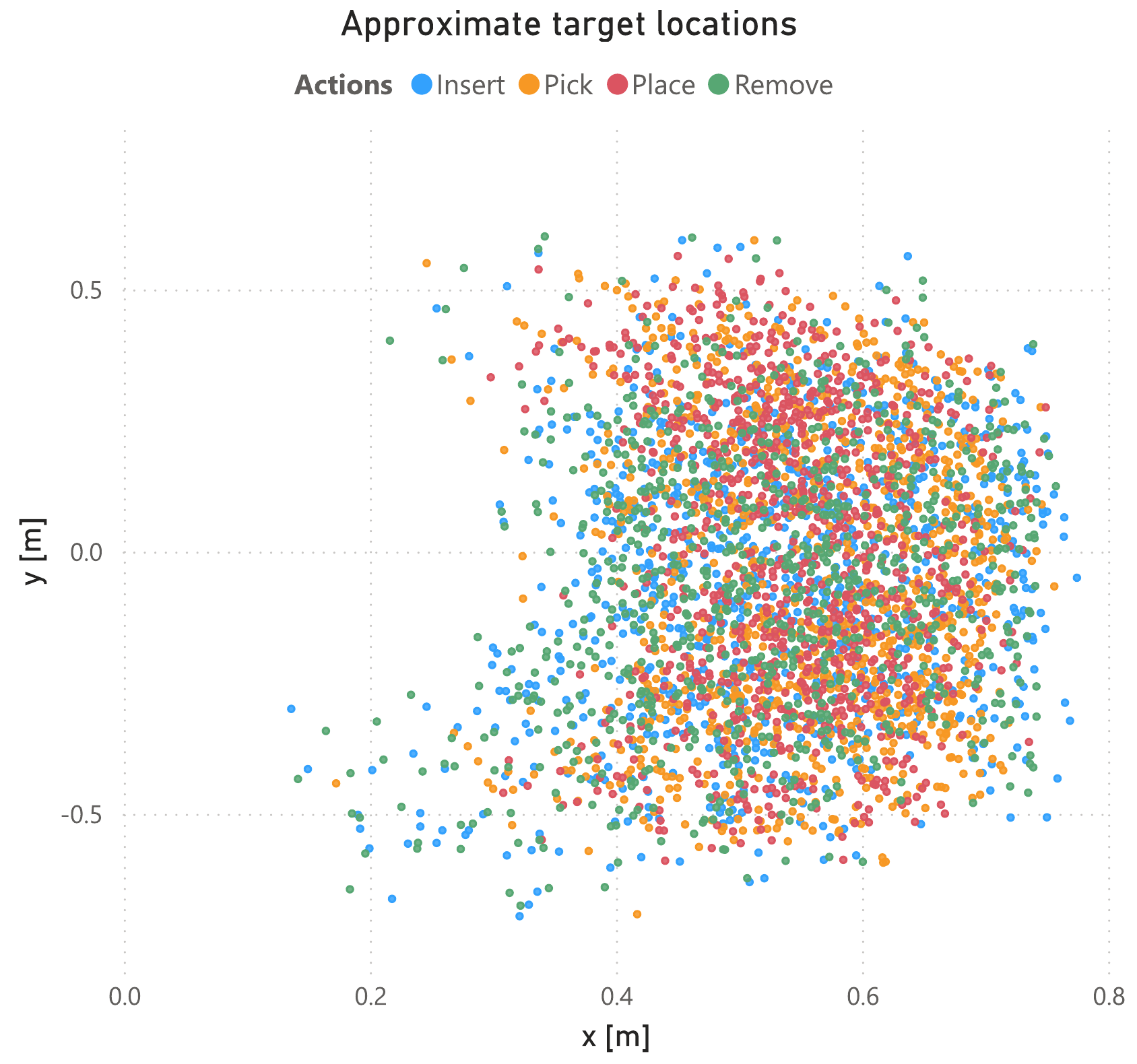}
    \caption{\textbf{Approximate interaction point for each of the actions.} The robot is placed at the point (0,0), facing the positive x-direction. REASSEMBLE has a large variety in the interaction points for each action.}
    \label{fig:stats_position}
\end{figure}

\subsection{Interaction point diversity}
Prior works have shown that diversity in the interaction points of actions within a dataset improves generalization and performance on downstream tasks~\cite{droid}. To ensure high diversity in the collected data, we instructed the operator to randomize the board and object poses for each trial during data collection. Figure~\ref{fig:stats_position} shows the approximate interaction point of all 4,551 demonstrations, obtained by sampling the final end-effector position of each demonstration. \textcolor{black}{The interaction points approximately show the positions of the objects and the board within the robot's workspace during data collection.} The robot is positioned at the origin and faces the positive x-axis direction. In most trials, the objects and the board were placed in front of the robot, within its workspace.

\begin{figure*}[htbp]
    \centering
    \begin{subfigure}[b]{\textwidth}
        \includegraphics[width=\textwidth]{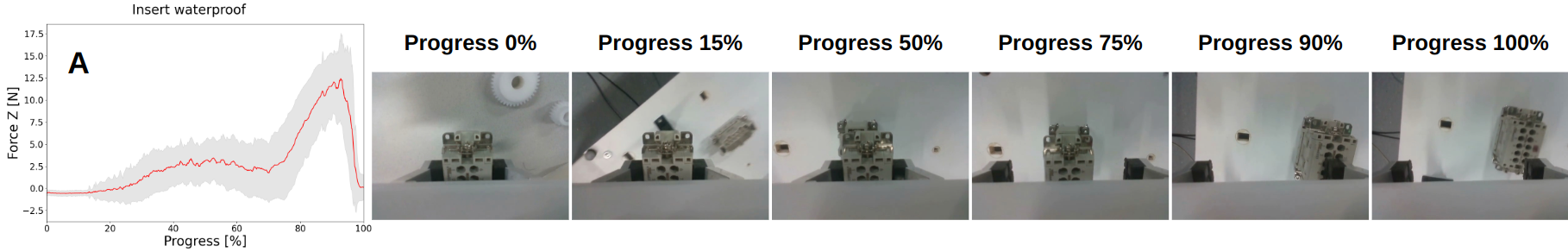}
    \end{subfigure}
    \begin{subfigure}[b]{\textwidth}
        \includegraphics[width=\textwidth]{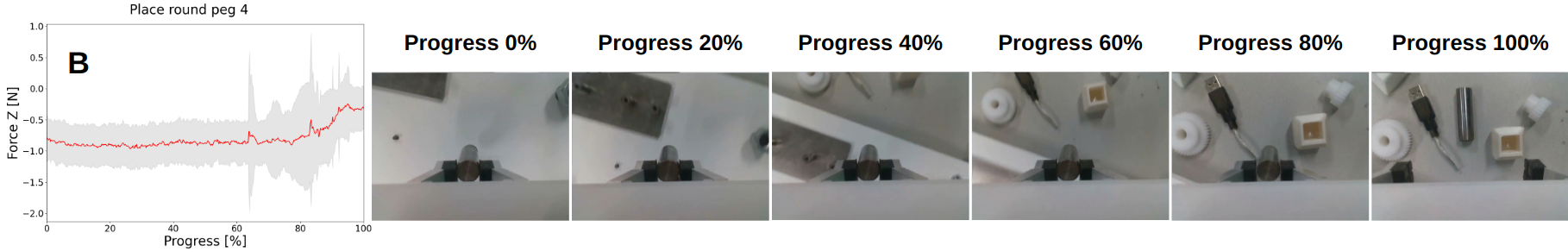}
    \end{subfigure}
    \begin{subfigure}[b]{\textwidth}
        \includegraphics[width=\textwidth]{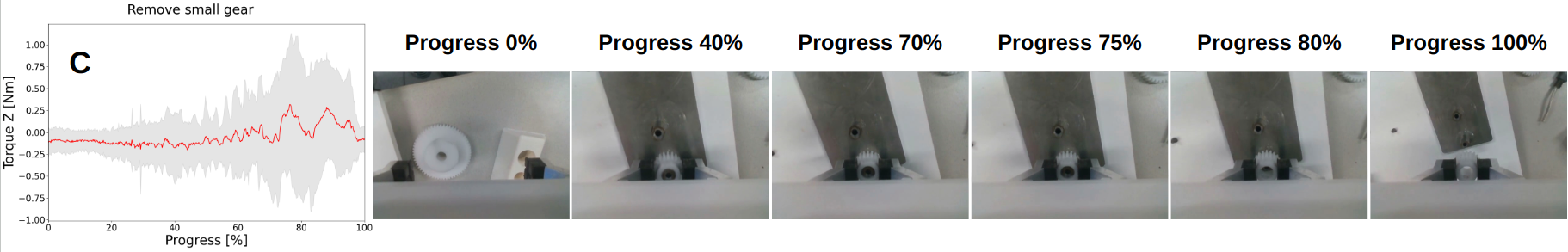}
    \end{subfigure}
    \caption{\textbf{Mean and standard deviation of force and torque measurements for selected actions.} To analyze the temporal evolution of each action, we normalize all demonstrations by their duration and compute the average wrench profile over the action progress. Distinct patterns emerge in the force and torque signals across actions, and similar trends are observed in other actions throughout the dataset.}
    \label{fig:stats_force}
\end{figure*}

\subsection{Force and Torque patterns}
We analyze whether any patterns occur in the demonstrated force and torque profiles of the actions. Since the demonstrations of each action have different durations, for visualization it is necessary to unify the number of the force and torque measurements to identify patterns across all demonstrations of a specific action. To achieve this, we normalize each demonstration by its duration, converting the demonstrations from the time domain to the "progress domain." A progress of 100\% indicates the action has finished. Next, we resample each demonstration to a common number of samples (500 in our case) by linearly interpolating the data. Finally, we plot the mean force and torque values along with their standard deviations for each action and visually search for meaningful patterns. The most interesting findings are presented in Figure~\ref{fig:stats_force}.

\textcolor{black}{
Figure~\ref{fig:stats_force}A shows the Z-component of the measured force for the \textit{Insert waterproof} action. Three distinct phases of the motion are visible: free-space movement from 0\% to 20\%, initial contact and alignment between 20\% and 75\%, and the pushing phase starting around 75\%. Around this point, the force briefly drops as the plug aligns with the socket and is released. From that moment until the end of the motion, the gripper continues to apply downward force until the connector snaps into place.
Figure~\ref{fig:stats_force}B shows the Z-component of the measured force for the \textit{Place round peg 4} action. Here, the effect of the object's weight is clearly observed. Around the 80\% mark, the gripper opens and releases the peg, which results in the force reading returning to nearly zero.
Figure~\ref{fig:stats_force}C shows the torque around the Z-axis for the \textit{Remove small gear} action. Periodic changes in torque are observed, which correspond to the operator twisting the gear back and forth while pulling it, in order to loosen and remove it.
}

\section{Benchmarks}
\subsection{Temporal Action Segmentation}

One of our goals when developing the REASSEMBLE dataset was to include multi-task annotations to enable the development of robotic algorithms for addressing challenges encountered at various stages of robotic system development. One such challenge, typically faced early in robot learning pipelines, is temporal action segmentation (TAS). The goal of TAS is to determine temporal boundaries between actions and label them in an untrimmed recording of a task demonstration. By segmenting a long-horizon demonstration into shorter, simpler actions, TAS simplifies policy learning for individual actions. Instead of learning one complex policy from the entire demonstration, more primitive policies can be learned and chained together.

More formally, TAS can be defined as follows: given a dataset of $N$ untrimmed task demonstrations $\mathcal{D} = \{d_i, s_i\}_{i=0}^N$, where $d_i$ represents a demonstration of varying length containing modalities such as vision, force, and/or robot proprioception information, and $s_i$ represents the ground truth segmentation, the goal is to learn a segmentation model $\mathcal{M}_\theta(d_i)$ such that $\mathcal{M}_\theta(d_i) = s_i$. In the computer vision community, TAS is typically posed as a supervised learning problem~\cite{liu2023diffusion}, where the model is trained using both data and ground truth action segments. In contrast, the robotics community often approaches TAS as an unsupervised problem~\cite{krishnan2017transition, willibald2022multi}, where patterns are learned only from demonstrations.

For benchmarking purposes, we evaluate the performance of a state-of-the-art visual TAS model, DiffAct~\cite{liu2023diffusion}. DiffAct leverages diffusion processes~\cite{ho2020denoising} to model and learn action boundaries. Diffusion processes work by progressively adding noise to the ground truth information and learning how to iteratively remove this noise. In conditional diffusion processes~\cite{dhariwal2021diffusion}, the denoising process is guided using additional information. For DiffAct, video visual features serve as the conditioning information. We follow the original DiffAct paper and use I3D video features~\cite{carreira2017quo}.

To extract I3D features, we preprocess RGB videos and optical flow data estimated using the RAFT~\cite{teed2020raft} model. We use only the wrist camera, as it provides the best perspective for distinguishing manipulated objects, being the closest camera to them during demonstrations. The video is downsampled from 30 to 10 frames per second to reduce computational time. We extract the visual features in windows of 21 frames, and with a stride of 1, resulting in one feature for each corresponding frame in the video.

In the TAS domain, five metrics are commonly used to evaluate and compare model performance:
\begin{itemize}
    \item \textbf{Frame-level accuracy:} Measures the proportion of correctly annotated frames relative to the total number of frames.
    \item \textbf{EDIT score:} Quantifies the number of label changes needed to convert the predicted segmentation into the ground truth.
    \item \textbf{F1 scores at 10\%, 25\%, and 50\% overlap:} Measure the precision and recall of action segments, with the overlap indicating the required temporal overlap between predicted and ground truth segments for them to be considered correct.
\end{itemize}
Among these, F1@50 is the most commonly used metric as it is the most stringent.

\begin{figure}[t]
    \centering
    \includegraphics[width=\linewidth]{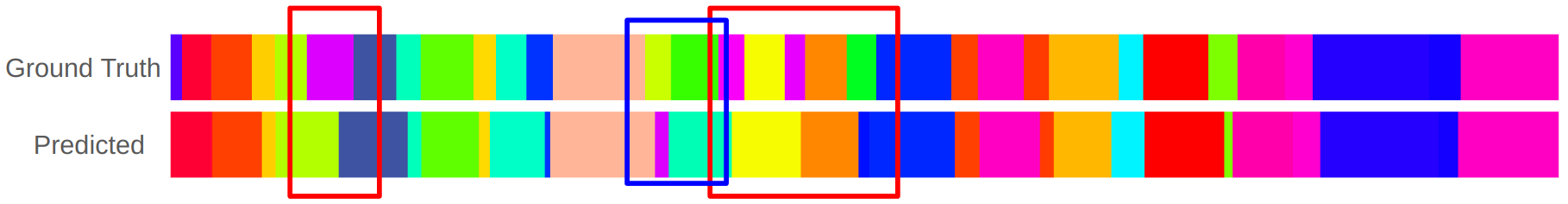}
    \caption{\textbf{Temporal action segmentation results.} In red, we highlight instances where the "Pick" action was not predicted by DiffAct. In blue, we mark areas where similar objects were confused. In this case, DiffAct confused "round peg 1" with "square peg 1" and "square peg 2."}
    \label{fig:tas}
\end{figure}

We use the default hyperparameter settings provided for the 50Salads dataset~\cite{50Salads}. The performance of DiffAct on the REASSEMBLE dataset is as follows: Accuracy 61.5\%, EDIT 47.8\%, F1@10 63.3\%, F1@25 58.4\%, and F1@50 44.1\%. For comparison, DiffAct achieves significantly higher performance on the 50Salads dataset: Accuracy 88.9\%, EDIT 85.0\%, F1@10 90.1\%, F1@25 89.2\%, and F1@50 83.7\%. 

We hypothesize that the lower performance of DiffAct on the REASSEMBLE dataset is due to the increased challenges it presents. Firstly, the REASSEMBLE dataset contains a significantly higher number of unique actions (69 compared to 19 in 50Salads), and many of the objects are less visually distinguishable from each other, such as gears and pegs. Additionally, the REASSEMBLE dataset has almost twice the median number of actions per video (36 compared to 19 in 50Salads). Furthermore, REASSEMBLE often includes sequences where very long actions (e.g., Insert and Remove) are separated by very short actions (e.g., Pick and Place). DiffAct frequently misses these short actions, as illustrated in Figure~\ref{fig:tas}. 

\textcolor{black}{One area that remains relatively unexplored is the effective fusion of multimodal data within TAS models, which has been limited partly due to a lack of suitable datasets. In that regard,  the REASSEMBLE dataset opens up new possibilities to explore various strategies for integrating multiple data modalities. We further investigated strategies for fusing multimodal data for robotic TAS models~\cite{sliwowski2025M2R2}. Preliminary results demonstrate improved performance through the integration of visual, auditory, force-torque (wrench), gripper, and pose information. These findings are promising, and we plan to conduct a more comprehensive analysis in future work to further validate and refine our approach.}

The REASSEMBLE dataset presents new opportunities for advancing temporal action segmentation (TAS) in robotics. One underexplored area is the proper fusion of multimodal data in TAS models, which has been limited partly due to a lack of suitable datasets. Additionally, the challenging nature of REASSEMBLE could facilitate research into embedding common-sense reasoning and hierarchical understanding within TAS models. For example, enabling models to learn that before inserting an object, it must first be picked up.

\begin{figure*}[htpb]
    \centering
    \includegraphics[width=\linewidth,]{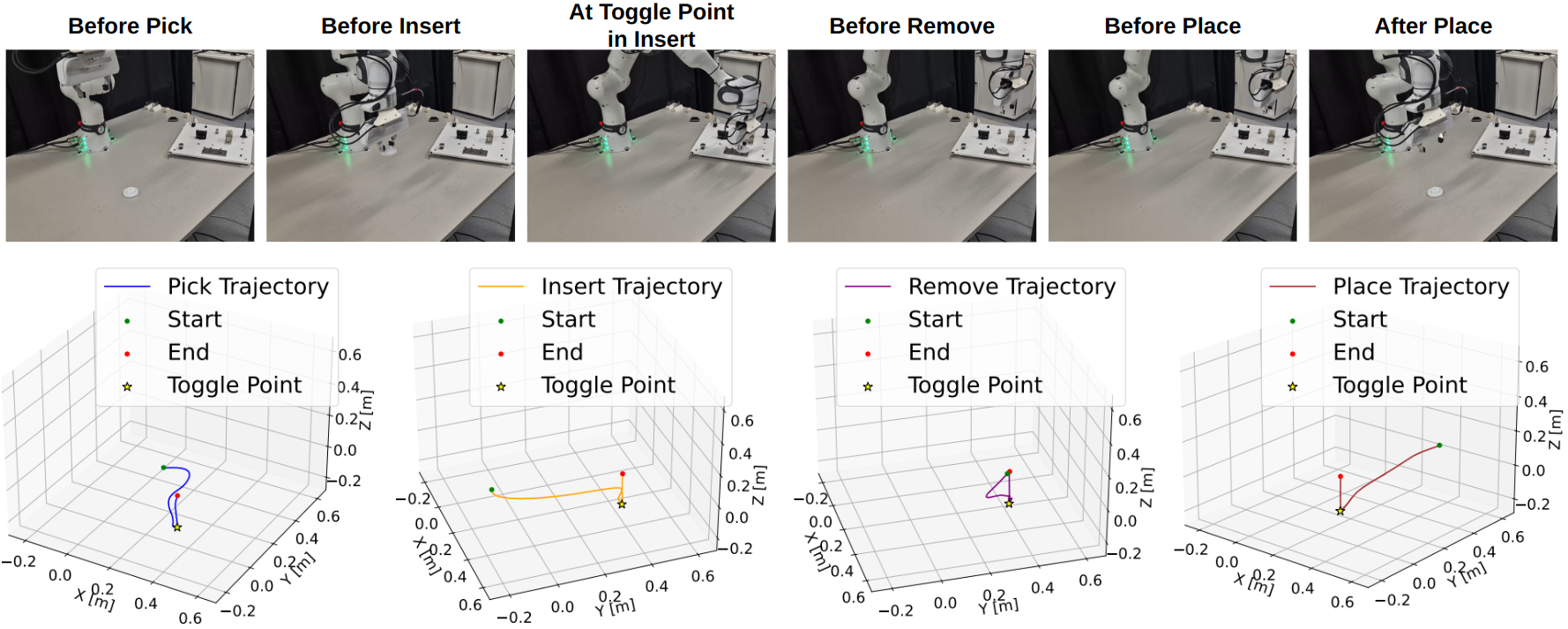}
    \caption{\textbf{Large Gear assembly \& disassembly} The figure illustrates the trajectories generated by the DMP framework for robotic assembly and disassembly of the large gear, including Pick, Insert, Remove, and Place motions. These trajectories are executed using an impedance controller. Key points such as the start, end, and toggle positions for gripper actions (specifically opening and closing) are highlighted, demonstrating the framework's ability to adapt trajectories based on updated object poses and task requirements.}
    \label{fig:dmp}
\end{figure*}

\subsection{Motion Policy Learning}
The primary objective of this study is to introduce a novel robot manipulation dataset specifically designed for contact-rich manipulation tasks, rather than to develop a new policy learning methodology. \textcolor{black}{We train and evaluate a language- and goal-conditioned diffusion model for MPL using the REASSEMBLE dataset. The model successfully captures motion profiles for gear assembly and disassembly, achieving a goal-point error of $\approx$1.3 cm, comparable to the SOTA method~\cite{chi2024universal}. While this error is sufficient for tasks like pick and place, it presents challenges for high-precision tasks such as insertion.} \textcolor{black}{The contemporary models such as diffusion policies and action chunking transformers have shown promising results across various tasks, they have yet to be optimized for high-precision applications such as assembly tasks involving the NIST board \cite{chi2023diffusion,zhao2023learning}.} Therefore, for the complete experimental evaluations, we adopt a well-established state-of-the-art approach by integrating Dynamic Movement Primitives (DMPs) into our framework \cite{saveriano2023dynamic,zhao2023robotic}. DMPs have been extensively researched over several decades, providing a robust mathematical foundation for learning and reproducing complex motion trajectories. 
In our approach, DMPs are utilized to learn the weights corresponding to specific motion conditions, particularly we focus on Cartesian positions.  The formulation of the DMPs in our system is governed by the following set of differential equations
\begin{align}
\tau_{\ddot{z}} &= \alpha_z \left( \beta_z (\mathbf{g} - \mathbf{y}) - \mathbf{z} \right) + \hat{\mathbf{f}}(\mathbf{x}), \\
\tau_{\dot{\mathbf{y}}} &= \mathbf{z}, \\
\tau_{\dot{\mathbf{x}}} &= \alpha_x \mathbf{x},
\end{align}
where \(\mathbf{y} \in \mathbb{R}^n\) represents the position, \(\mathbf{z} \in \mathbb{R}^n\) is the velocity, \(\mathbf{x} \in \mathbb{R}^n\) is the canonical system state, \(\mathbf{g} \in \mathbb{R}^n\) denotes the goal position, and \(\alpha_z, \beta_z, \alpha_x \in \mathbb{R}\) are gain coefficients that govern the system's convergence and stability properties. The nonlinear forcing term \(\hat{\mathbf{f}}(\mathbf{x})\) is critical for learning complex trajectories and is defined as
\begin{align}
\mathbf{f}(\mathbf{x}) = \frac{\sum_{i=1}^{N} \mathbf{w}_i \psi_i(\mathbf{x})}{\sum_{i=1}^{N} \psi_i(\mathbf{x})} \mathbf{x},  
\end{align} where \(\mathbf{w}_i \in \mathbb{R}^n\) are the learned weights, and \(\psi_i(\mathbf{x})\) are the basis functions typically chosen as Gaussian functions
\begin{align}
\psi_i(\mathbf{x}) = \exp\left( -h_i (\mathbf{x} - c_i)^2 \right),
\end{align} with \(h_i \in \mathbb{R}\) representing the width parameters and \(c_i \in \mathbb{R}\) the centers of the Gaussian functions. The weights \(\mathbf{w}_i\) are learned from demonstration data by minimizing the error between the demonstrated trajectories and those generated by the DMP framework, often using least squares optimization techniques.

During inference, DMP framework generates motion trajectories by adapting to updated start and goal points. The start point \(\mathbf{y}_0\) is extracted from the robot's current pose at the time of inference, while the goal pose \(\mathbf{g}\) is determined using an RGB-D-based object state estimation pipeline. This pipeline employs a YOLO model trained to detect NIST objects, enabling accurate real-time object localization via an RGB-D camera. Hand-eye calibration is performed to accurately map detected object positions to the robot's coordinate frame. For non-symmetric objects such as the waterproof connector and D-SUB connector, Principal Component Analysis (PCA) is applied to the point cloud data captured by the RGB-D camera. PCA identifies the primary axes of variance in the data, which can be used to determine the object's orientation from the point cloud data of the associated object.

The effectiveness of the proposed framework is evaluated in large gear assembly and disassembly tasks serving as a representative test case. Once the DMP generates a trajectory based on the learned weights and the provided start and goal points, the robot executes the trajectory using an impedance controller. This controller ensures compliance and adaptability to environmental uncertainties, as depicted in Figure \ref{fig:dmp}. Gripper actions are conditioned on the goal error, such that upon reaching the goal pose, gripper operations are executed according to task-specific conditions.

Performance demonstrations highlight successful object detection and manipulation tasks, including the challenging assembly and disassembly of gears. As shown in Figure \ref{fig:dmp}, the DMP framework effectively learns the required motion trajectories for these tasks, facilitating precise operations such as picking, inserting, removing, and placing the large gear. For insertion tasks, a spiral motion search is conducted on the \(xy\)-plane while maintaining a constant downward force along the \(z\)-axis \cite{chhatpar2001search}. This approach utilizes force thresholds learned from the dataset to optimize insertion precision. The spiral search strategy balances the trade-off between compliance and accuracy, ensuring successful gear insertion. By integrating DMPs with our perception pipeline and control strategies, the proposed framework achieves robust, high-precision performance in contact-rich manipulation tasks. We conducted a series of experiments to evaluate the performance of a large gear assembly and disassembly task, encompassing pick, insert, remove, and place motions. Each action was executed 10 times to assess its success rate and failure modes. The pick action was successfully executed in 8 out of 10 trials, with failures occurring due to the gear slipping from the gripper. The insert motion exhibited the lowest success rate, similar to human performance, with 7 out of 10 successful insertions. The primary failure mode for the insertion task occurred when the spiral search trajectory completed, but the gear remained outside the target position. The remove action was successfully performed in 8 out of 10 trials, with failures arising when the gripper mistakenly grasped the gear holding plate instead of the gear. This issue was attributed to tracking errors in the impedance controller, which could potentially be mitigated by increasing impedance gains. The place motion was successful in all trials. These results highlight the reliability and adaptability of DMPs and force thresholds learned from the dataset in object assembly and disassembly scenarios, demonstrating their effectiveness in executing complex manipulation tasks. 

\subsection{Task Execution and Monitoring using REASSEMBLE}
When deploying learned policies, task execution can fail due to various reasons. First, the learned motion policies might not generalize properly to new environmental settings. Second, perception systems can introduce errors in object positioning, leading to incorrect task parameters being passed to motion planners. Third, errors may arise from controllers or robot kinematic limits. Finally, humans may interrupt robots during task execution. Therefore, detecting such errors and recovering from them is crucial.

To demonstrate that the annotations within the REASSEMBLE dataset can be used to effectively learn anomaly detection pipelines, we train and develop an execution monitoring pipeline similar to the one presented in the ConditionNET paper~\cite{sliwowski2024conditionnet}. To ensure a reliable anomaly detection framework, we train ConditionNET with observations from the wrist-mounted camera, as objects are easiest to distinguish from this view. Experimentally, we define the range where preconditions are satisfied between the 10\% and 20\% mark of the ground truth action segmentation label, and effects between the 95\% and 100\% mark. Based on this definition, we generate the ground truth labels used to train the ConditionNET model. We evaluate the anomaly detection performance using three metrics: Accuracy, Recall, and Precision. ConditionNET achieves an accuracy, precision, and recall of 96\%. For motion policies, we use the DMPs learned for the picking task described in the previous section. The behavior of the robot and its reactions to anomalies are expressed using a behavior tree~\cite{BT}, similar to the one presented in~\cite{sliwowski2024conditionnet}.

\begin{figure}
    \centering
    \includegraphics[width=\linewidth]{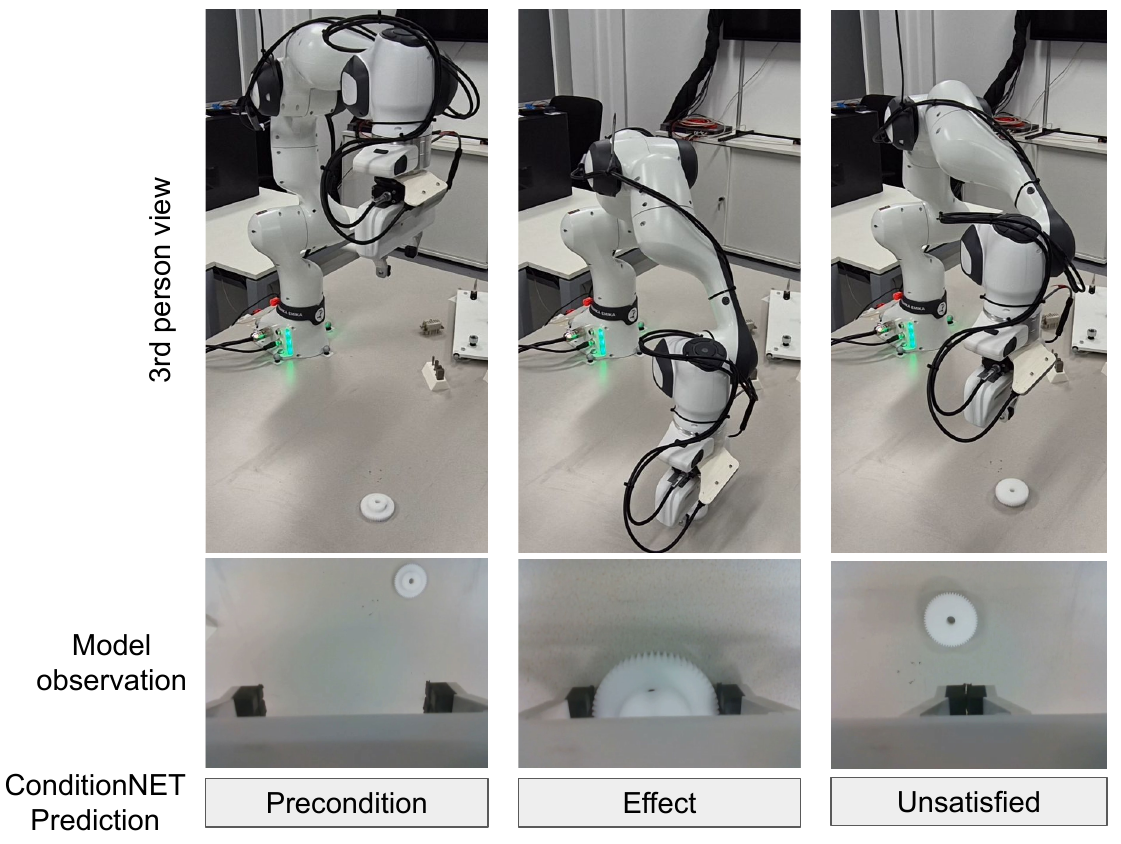}
    \caption{\textbf{Example snapshots from the execution monitoring experiments.} On the left, the observation just before starting the execution of the approach primitive. In the middle, the moment after the gear is grasped and before the lift primitive is executed. On the right, the situation after the gear falls out of the gripper.}
    \label{fig:EM}
\end{figure}

Figure~\ref{fig:EM} showcases example situations that occur during the execution of the picking task. The leftmost figure represents the initial observation before executing the picking motion. In this case, we observe that ConditionNET predicts the state to be the precondition state of the action, as the gripper is empty. The middle figure shows the situation after the approach motion has been executed and the gear has been grasped. Here, ConditionNET predicts that the effects have been met. The last figure illustrates the case where the gear has fallen out of the gripper. ConditionNET predicts "Unsatisfied," as the gripper is closed and no object is being held. For a more detailed overview of the execution monitoring experiments, please refer to the supplementary video.

\section{Limitations}
\textcolor{black}{The REASSEMBLE dataset leverages the NIST Task Board\#1, which consists of geometrically similar objects with limited visual features. Objects such as round and square pegs and gears pose a particular challenge, as they share similar visual characteristics but differ proprioceptively. This suggests that proprioceptive data may be more effective than visual data for object identification in temporal action segmentation tasks. While this limitation stems from the design of the selected NIST Assembly Task Board, we plan to expand the dataset with additional boards. In particular, Task Board\#4 features objects, such as cables, that are more similar in terms of proprioceptive properties but can be distinguished visually.
Additionally, the dataset not including depth information constitutes a limitation for robotic perception and manipulation. It was excluded as with the depth data, the size of the dataset would become approximately 12 times larger. Depth data provides three-dimensional spatial context that enables robots to accurately perceive object geometries, estimate distances, and plan precise manipulation trajectories. However, current approaches are developing manipulation methods using only RGB images \cite{chi2023diffusion,brohan2022rt}.}

\section{Conclusion}
In this work, we present the REASSEMBLE dataset, designed for long-horizon and contact-rich assembly and disassembly manipulation tasks. The REASSEMBLE dataset leverages the standardized and well-established NIST Task Board \#1, which focuses on tasks such as gear meshing and insertion of pegs and electrical connectors. We collected a total of 4,551 action demonstrations, of which 4,035 were successful. The dataset includes data from multiple modalities, such as multi-view RGB cameras, robot proprioception, force and torque sensors, microphones, and event cameras. Furthermore, we provide multi-task annotations and benchmark results for key tasks, including temporal action segmentation, motion policy learning, and anomaly detection.

The REASSEMBLE dataset addresses the scarcity of long-horizon and contact-rich manipulation datasets in robotics. We aim to facilitate research on robust and generalizable contact-rich manipulation agents capable of learning both assembly and disassembly tasks. We hope our initial results will encourage the robotics community to collaborate on solving the challenges presented by this dataset and advance the field of contact-rich manipulation.

\section*{Acknowledgments}
This work has been partially supported by the European Union project INVERSE under grant agreement No. 101136067, and in part by the Robot Industry Core Technology Development Program under Grant 00416440 funded by the Korea Ministry of Trade, Industry and Energy (MOTIE).

\bibliographystyle{IEEEtran}
\bibliography{main}

\begin{appendices}

\section{Data visualization tool}
\label{sec:appenda}
\textcolor{black}{
To simplify viewing the dataset, we designed a visualization tool based on the {\tt rerun} Python package. {\tt rerun} facilitates the easy visualization of various data types stored in the REASSEMBLE dataset and ensures that the data is displayed according to the recorded timestamps. }

\textcolor{black}{
To visualize the event camera data, we follow the standard practice of accumulating event information over 33 ms windows to generate a 30 FPS video. The timestamp of each frame in the generated video corresponds to the timestamp of the last chunk of event data included in the window. Since {\tt rerun} does not support audio playback, we instead display the waveform of the recorded audio signals.}

\begin{figure*}[htpb]
    \centering
    \includegraphics[width=\linewidth]{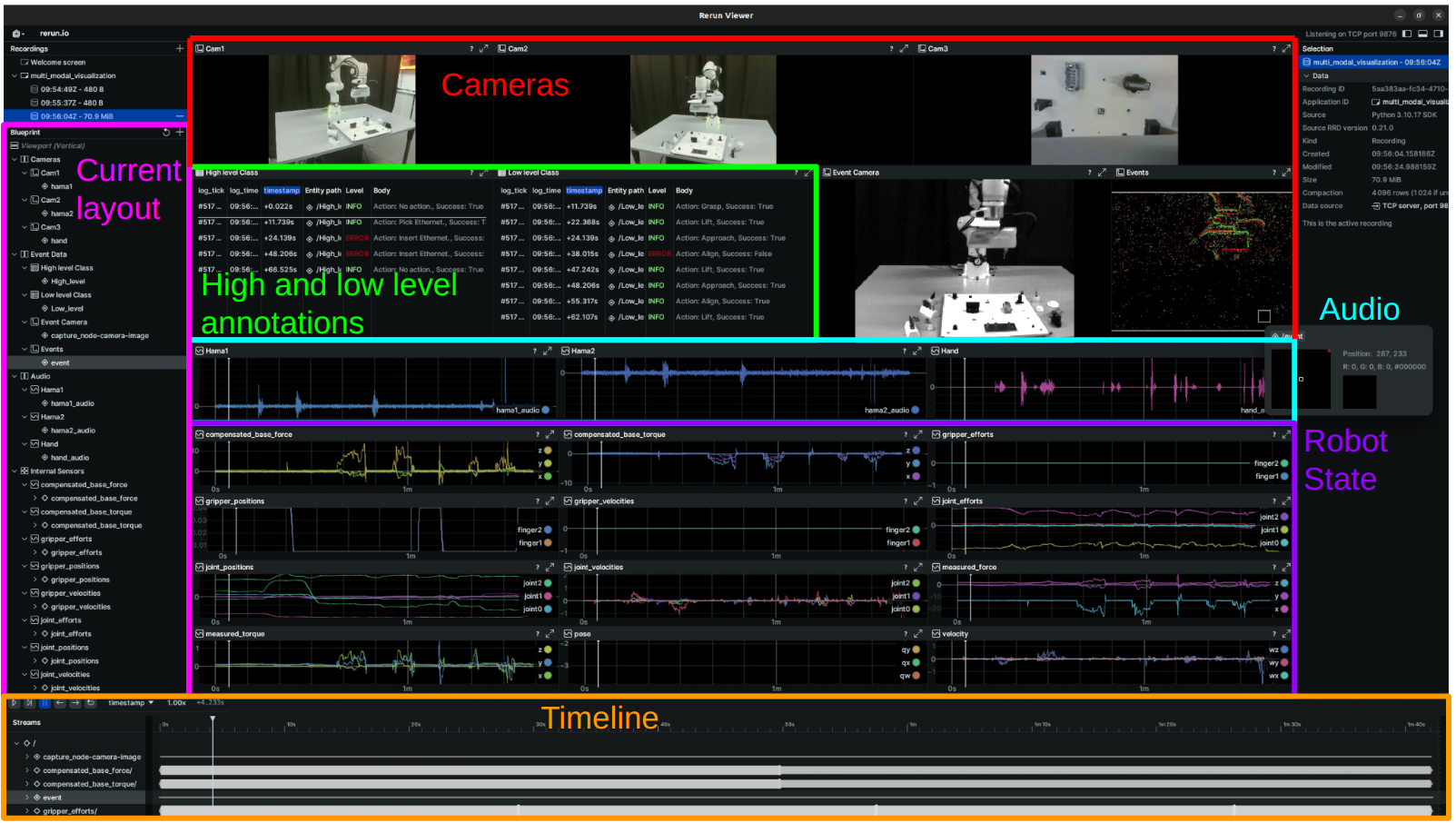}
    \caption{Snapshot from the dataset visualization tool. Best viewed in color.}
    \label{fig:visualizer}
\end{figure*}

\textcolor{black}{
Figure~\ref{fig:visualizer} showcases the data visualization tool. Green-marked widgets display all action segments and their success annotations, with the current action underlined. Red widgets show external and wrist camera views, while the bottom-right widget visualizes event data. Blue widgets illustrate audio waveforms from all microphones, and purple ones present proprioceptive data, including compensated forces and torques, joint and gripper states, and end-effector pose and orientation. The orange-marked timeline at the bottom displays the exact timestamps for each sensory modality. The rightmost widget shows the layout structure (Blueprint) of the visualization tool. The tool is accessible online via the \href{https://dsliwowski1.github.io/REASSEMBLE_page/}{project website}.
}

\begin{figure}
    \centering
    \includegraphics[width=\linewidth]{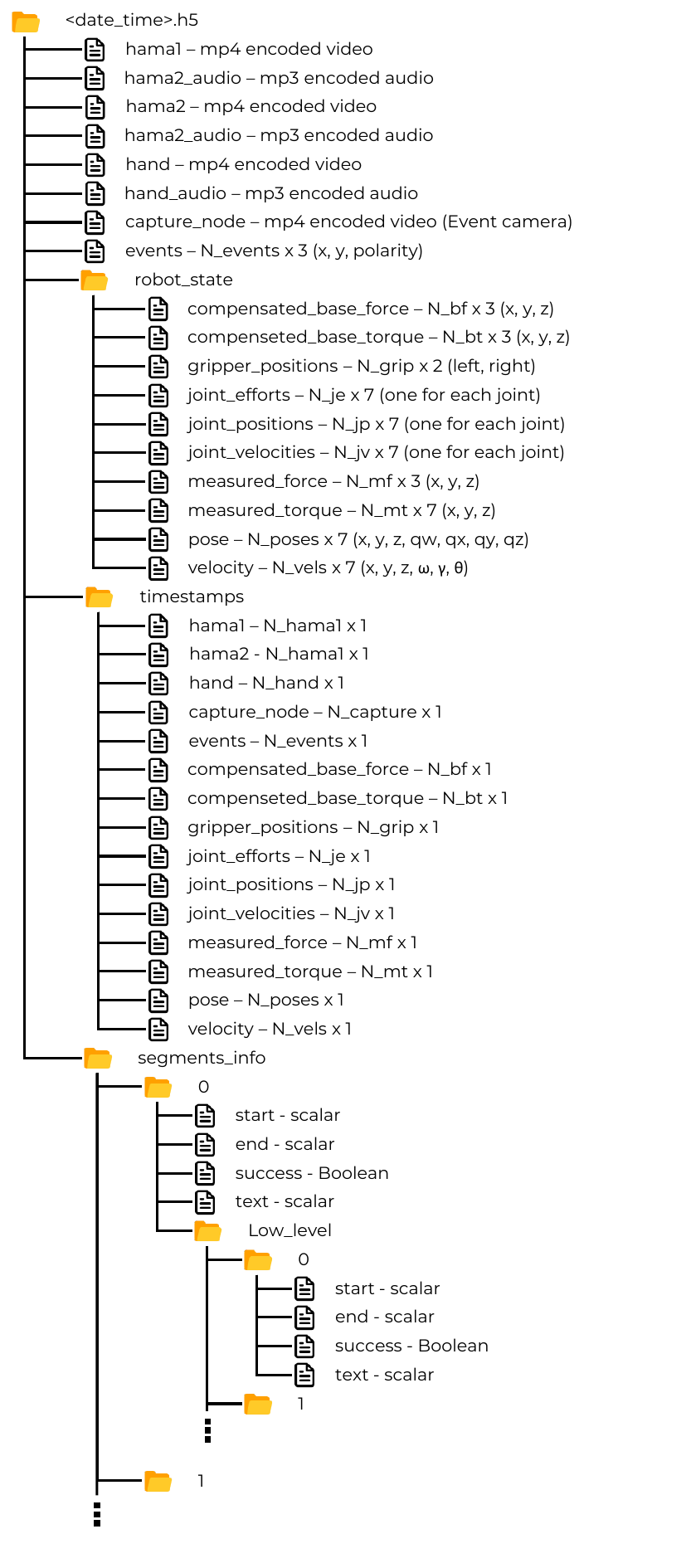}
    \caption{\textbf{Structure of a single h5 file.} Each trial is recorded into a separate h5 file. We save the data at their native publishing frequency, allowing for task-specific data synchronization and sampling for downstream tasks. To ensure easy synchronization of all messages, we save their published timestamps in the file.}
    \label{fig:h5structure}
\end{figure}

\section{Dataset structure}
\label{sec:appendB}
\textcolor{black}{Figure~\ref{fig:h5structure} shows the structure of a single demonstration file in the REASSEMBLE dataset. Each demonstration is stored in an individual file, identified by the date and time the demonstration began. The HDF5 (.h5) format distinguishes between two types of structures: \textit{datasets}, which contain the raw data, and \textit{groups}, which are folder-like elements that can contain multiple datasets or subgroups. In Figure~\ref{fig:h5structure}, groups are represented by folder icons, and datasets by file icons. Video, audio, and event data are stored directly within the main group of the file. Video and audio are stored as encoded byte strings to reduce memory usage, while event data is stored as arrays. The robot’s proprioceptive data is stored in the \texttt{robot\_state} group as arrays. Since each sensor modality is recorded at its own measurement frequency, the arrays have different lengths. To synchronize the sensory information, each modality’s timestamps are stored in the \texttt{timestamps} group. Action segment information is stored in the \texttt{segments\_info} group. Each segment has its own subgroup, identified by its order in the demonstration. Each segment contains a starting timestamp, ending timestamp, a success flag, and a natural language description of the action. Skills are stored within a \texttt{low\_level} subgroup inside each segment, and follow the same structure as the high-level annotations. Additionally, each HDF5 file has an associated JSON file containing the poses (positions and quaternions) of the cameras and the board. Code for loading and saving the HDF5 files is available on \href{https://github.com/TUWIEN-ASL/REASSEMBLE}{GitHub}.
}


\begin{figure*}[htbp]
    \centering
    \begin{subfigure}[b]{0.24\textwidth}
        \includegraphics[width=\textwidth]{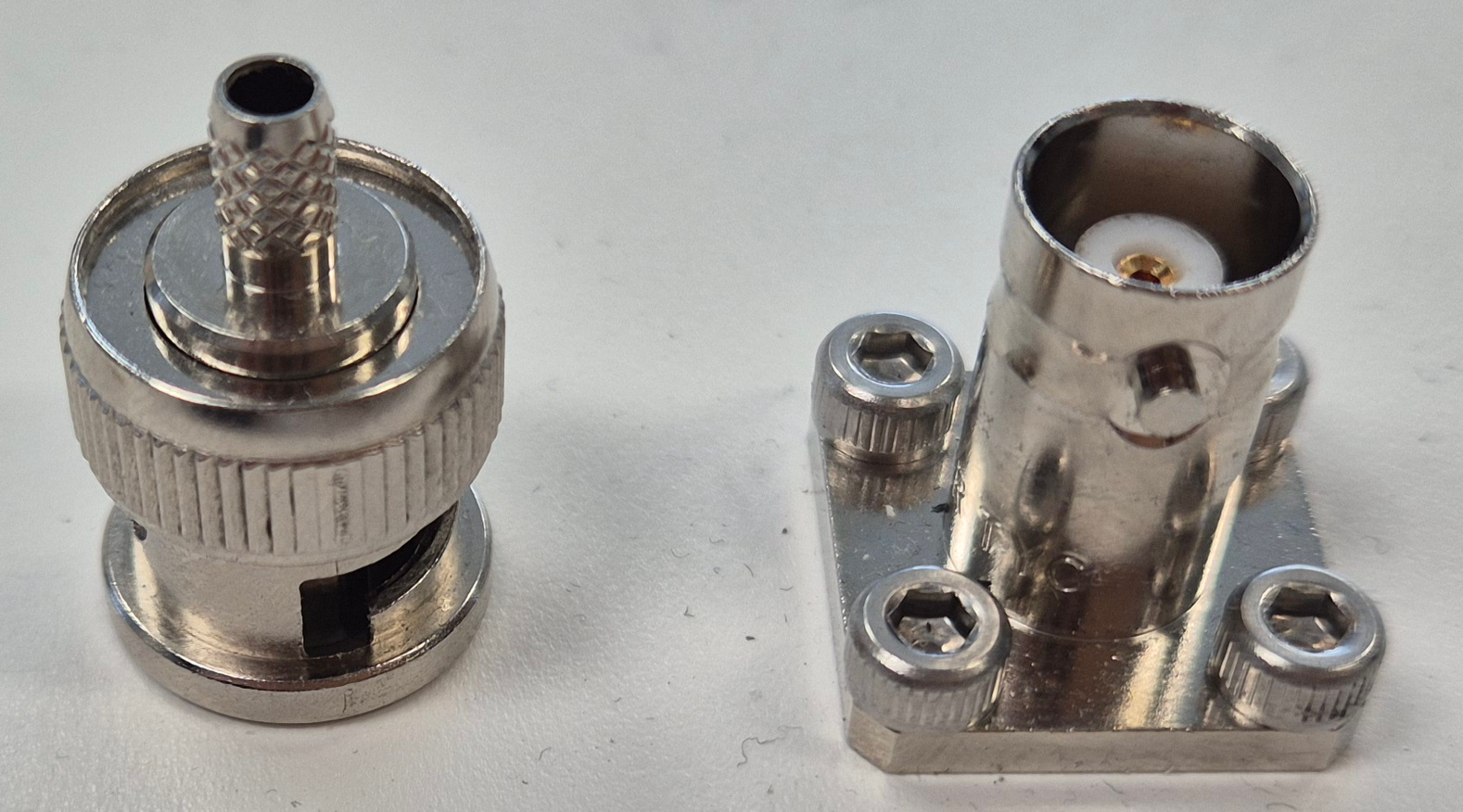}
        \caption{BNC connector and socket. Approx. diameter: 15\,mm.}
    \end{subfigure}
    \begin{subfigure}[b]{0.24\textwidth}
        \includegraphics[width=\textwidth]{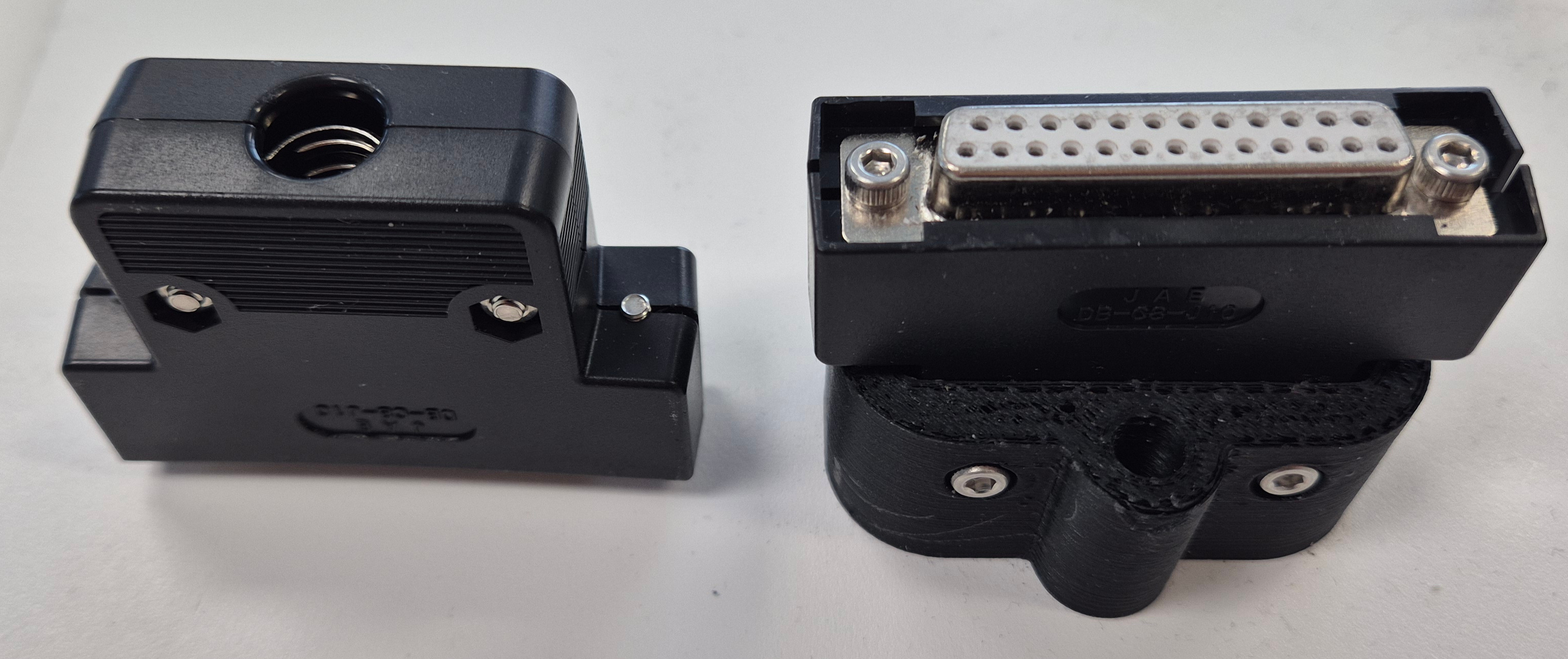}
        \caption{D-SUB connector and socket. Approx. size: 60$\times$20\,mm.}
    \end{subfigure}
    \begin{subfigure}[b]{0.24\textwidth}
        \includegraphics[width=\textwidth]{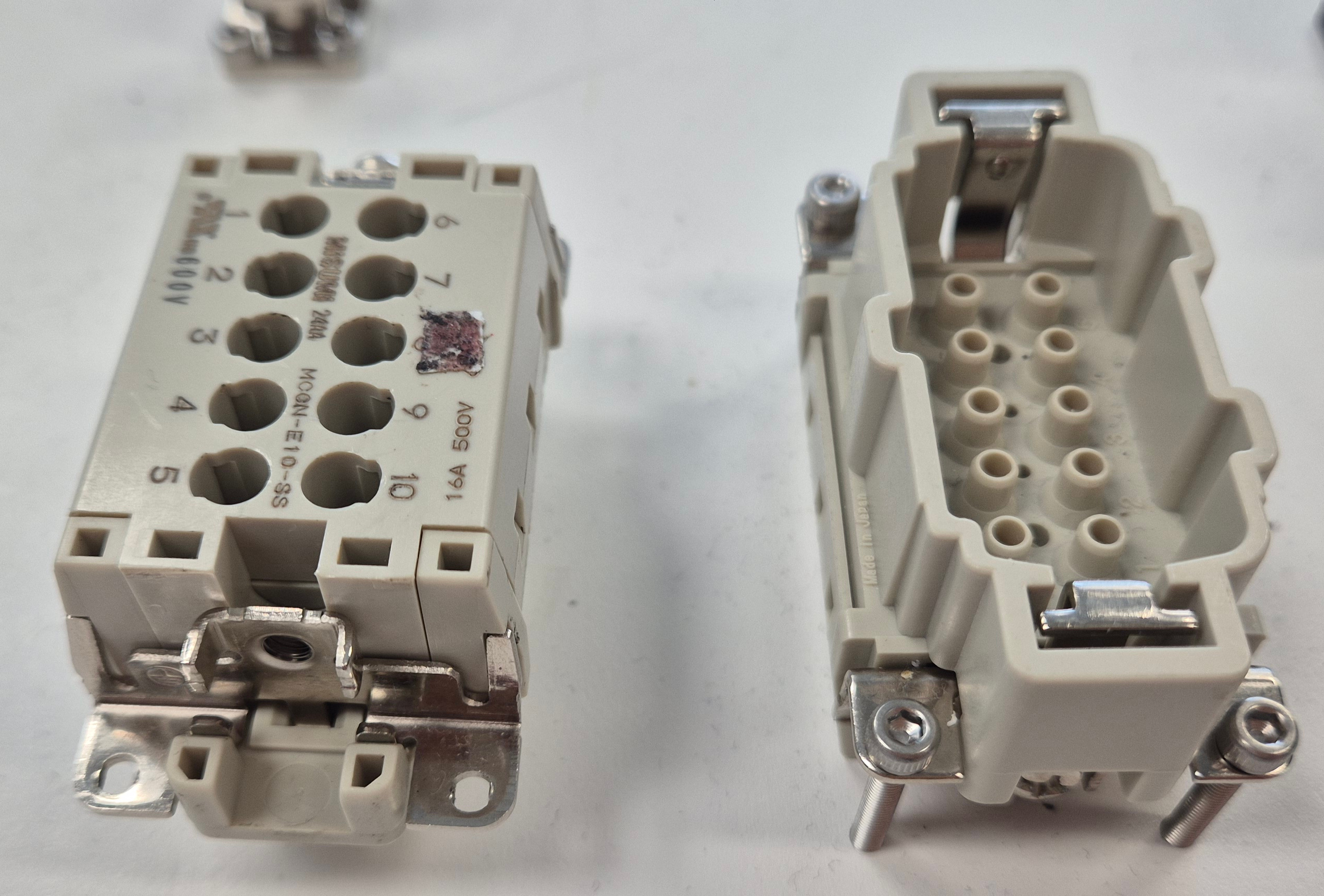}
        \caption{Waterproof connector and socket. Approx. size: 50$\times$35\,mm.}
    \end{subfigure}
    \begin{subfigure}[b]{0.24\textwidth}
        \includegraphics[width=\textwidth]{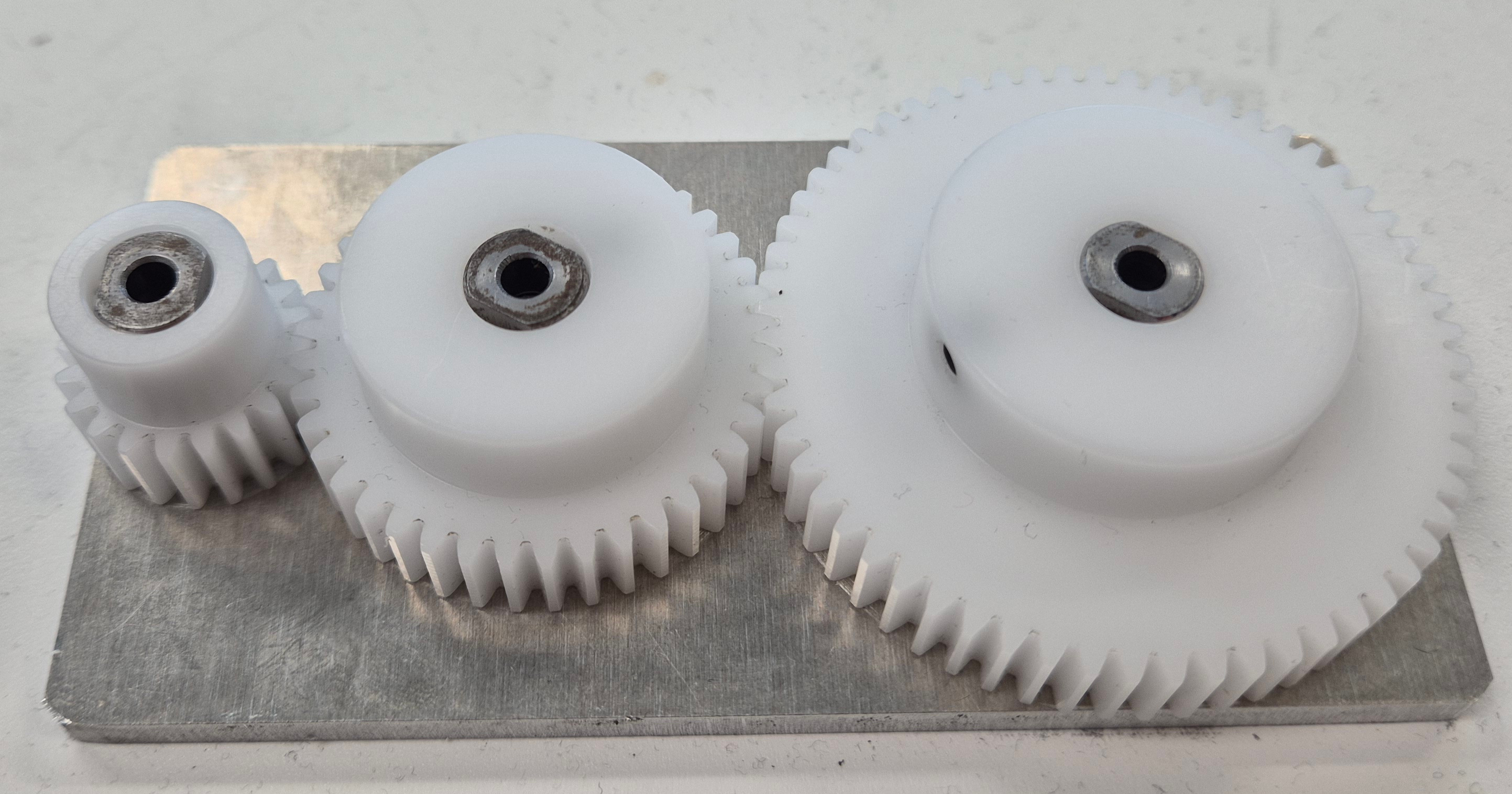}
        \caption{Gears: small, medium, and large (from left to right), mounted on a plate. Approx. diameters: 20\,mm, 40\,mm, and 60\,mm.}
    \end{subfigure}
    \begin{subfigure}[b]{0.24\textwidth}
        \includegraphics[width=\textwidth]{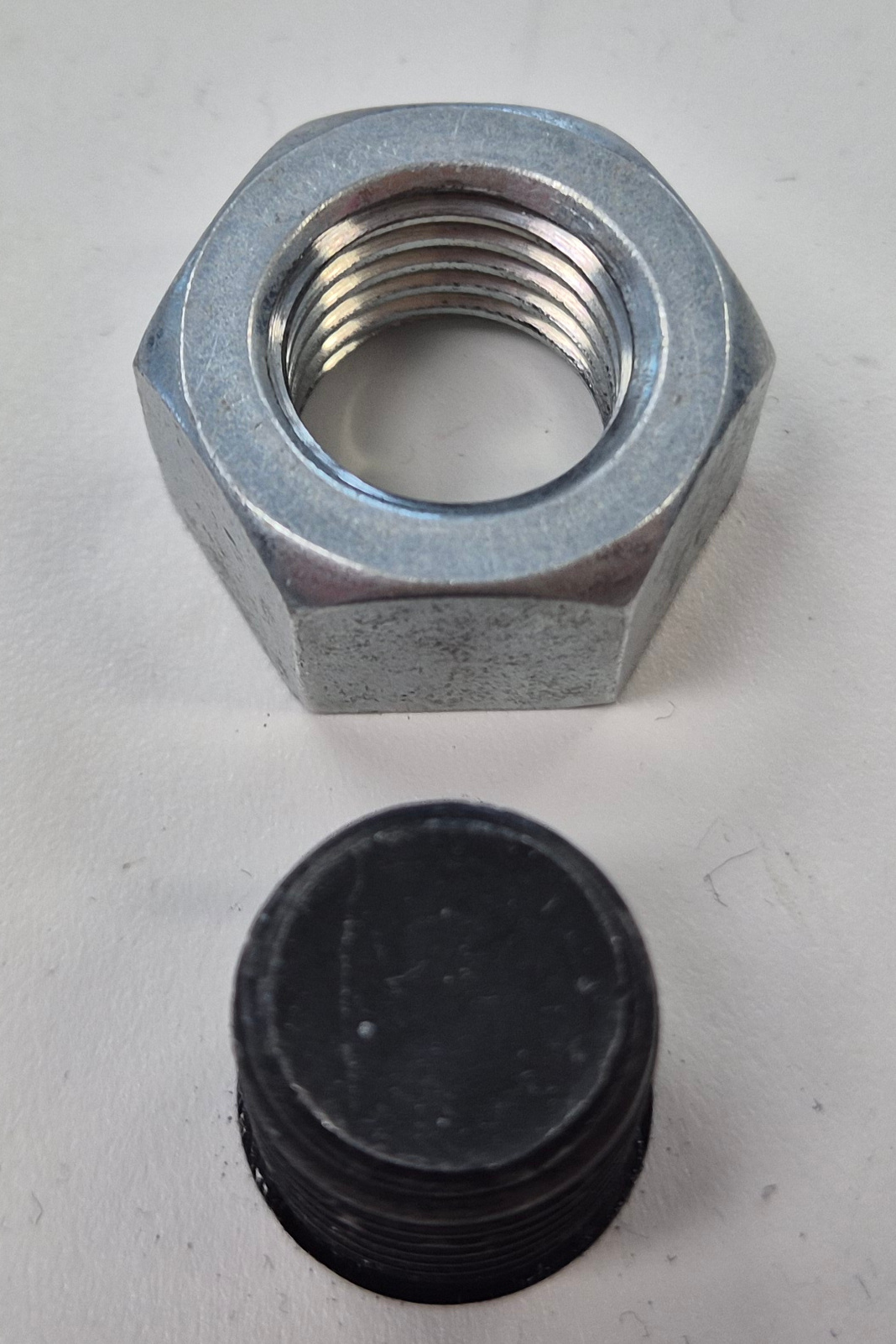}
        \caption{Nut 4 and its socket. Approx. diameter: 25\,mm.}
    \end{subfigure}
    \begin{subfigure}[b]{0.24\textwidth}
        \includegraphics[width=\textwidth]{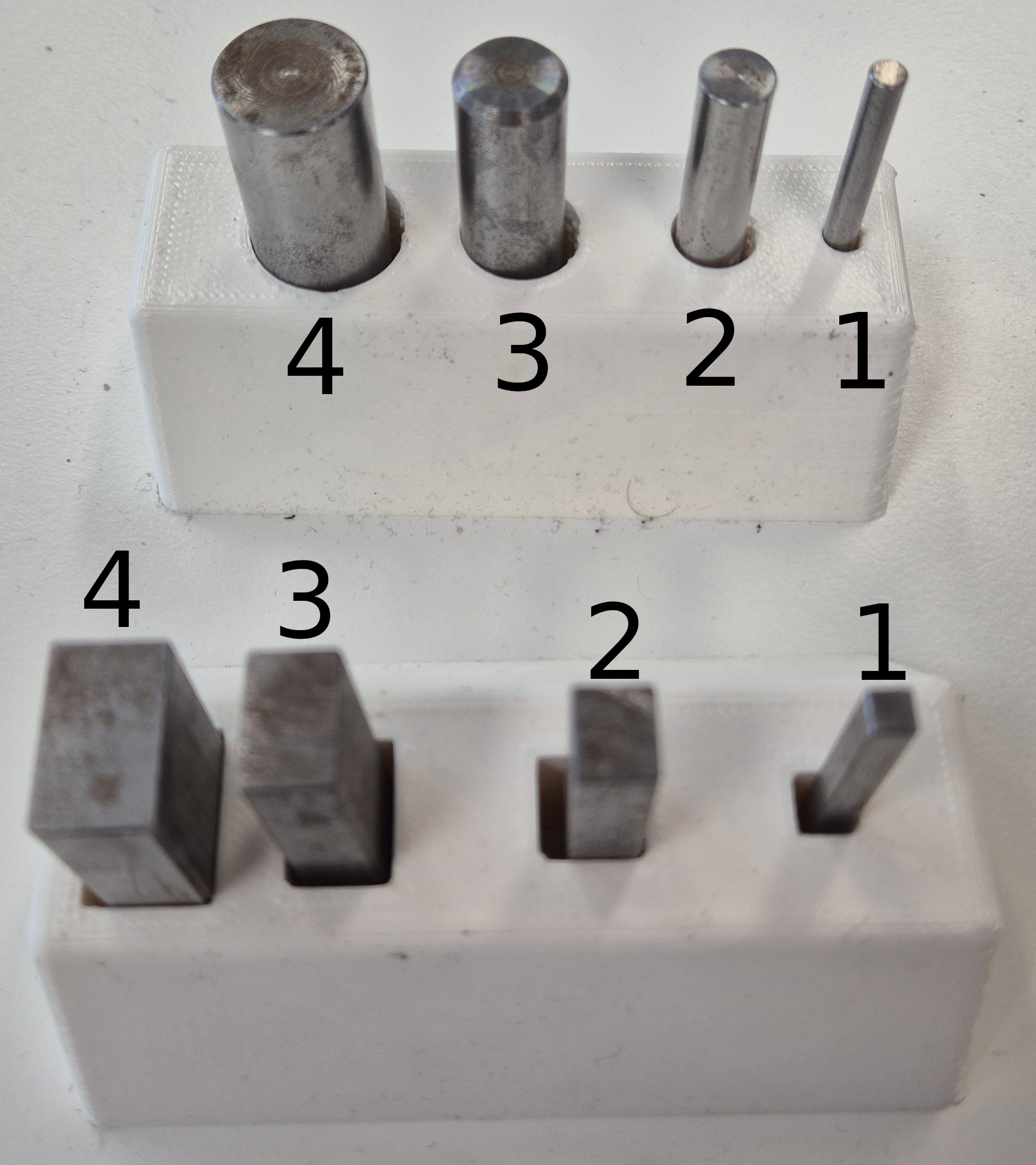}
        \caption{Top: round pegs. Bottom: square pegs. Round peg diameters (sizes 1–4): 4\,mm, 8\,mm, 12\,mm, 16\,mm. Square peg dimensions (sizes 1–4): 4\,mm, 8$\times$7\,mm, 12$\times$8\,mm, 16$\times$10\,mm.}
    \end{subfigure}
    \begin{subfigure}[b]{0.24\textwidth}
        \includegraphics[width=\textwidth]{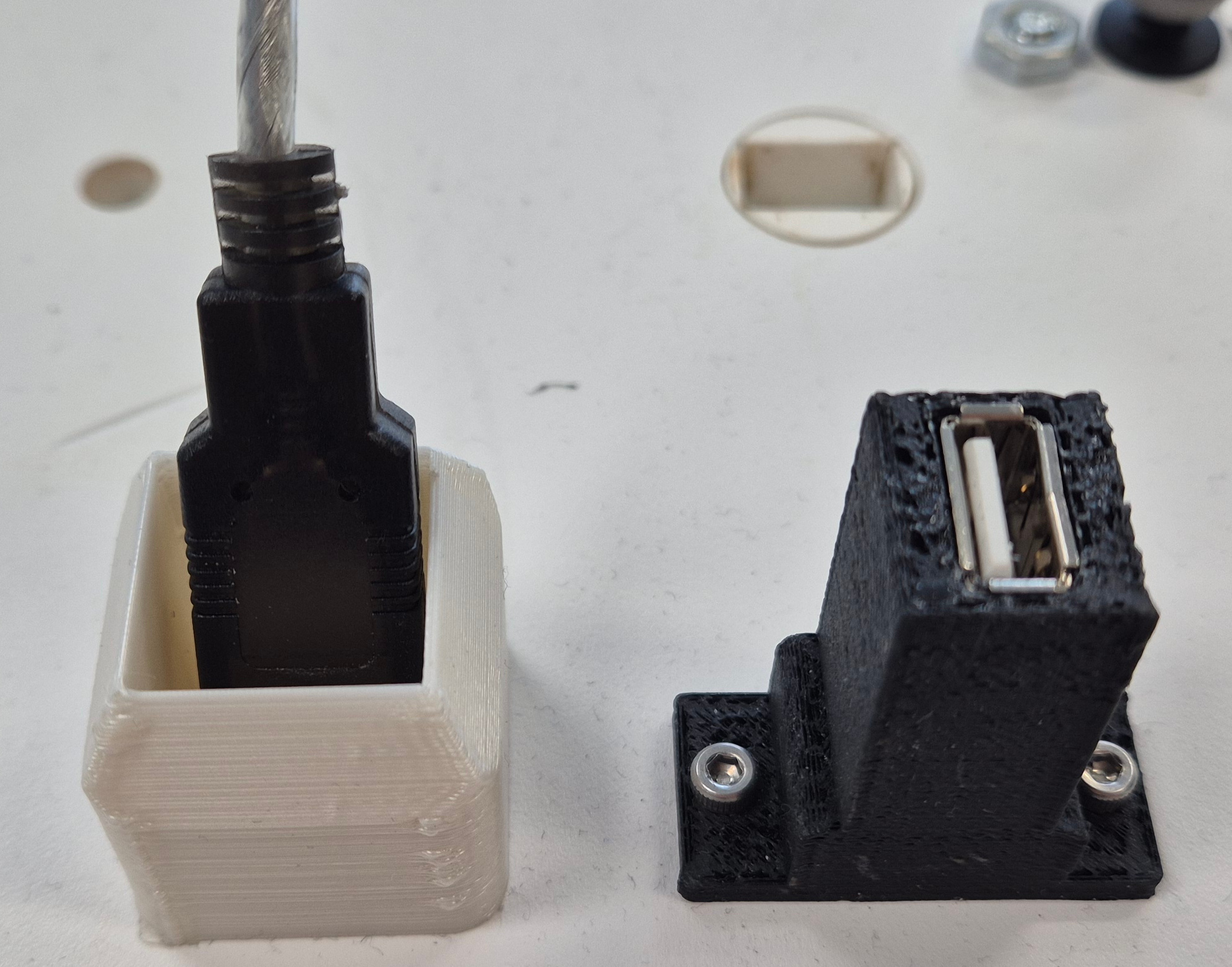}
        \caption{USB connector and socket. The connector is held in a custom holder.}
    \end{subfigure}
    \begin{subfigure}[b]{0.24\textwidth}
        \includegraphics[width=\textwidth]{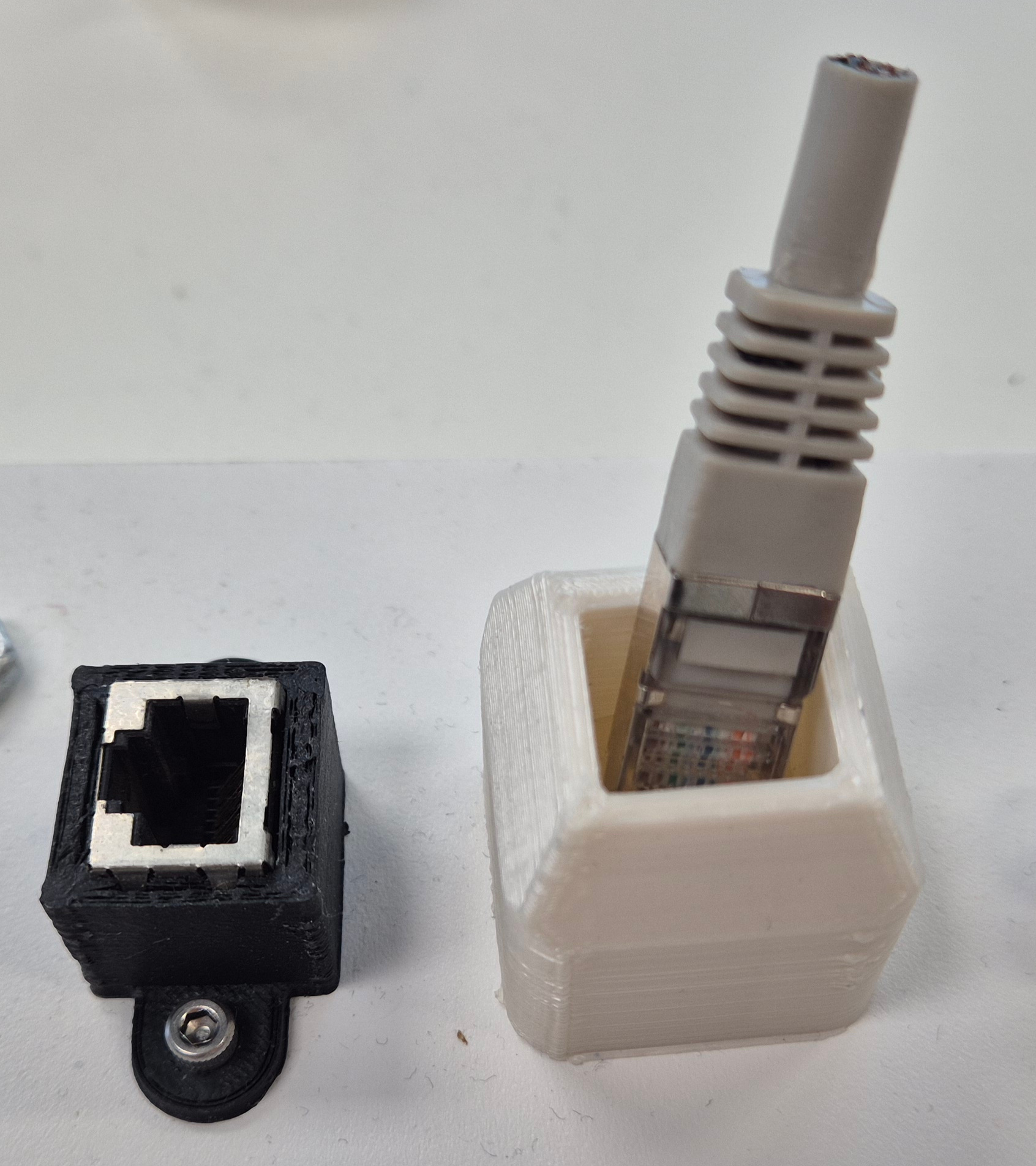}
        \caption{Ethernet connector and socket. The connector is held in a custom holder.}
    \end{subfigure}
    \caption{Objects and connectors used in the REASSEMBLE dataset.}
    \label{fig:objects}
\end{figure*}

\section{Objects}
\label{sec:appendD}
\textcolor{black}{
Figure~\ref{fig:objects} shows photos of the objects, connectors, and holders used in the REASSEMBLE dataset. We use 17 objects from the NIST task board 1 in the REASSEMBLE dataset:
\begin{itemize}
    \item 5 electrical connctors: Ethernet, USB, waterproof, BNC, D-SUB
    \item 3 gears: small, mediun, and large
    \item 1 M16 Nut
    \item 4 round pegs: 4mm, 8mm, 12mm, and 16mm
    \item 4 square pegs: 4mm x 4mm, 8x7, 12x8, 16x10
\end{itemize}
}

\end{appendices}

\end{document}